\documentclass[10pt,twocolumn,letterpaper]{article}

\usepackage[pagenumbers]{cvpr} 

\usepackage{multirow}
\usepackage{makecell}
\usepackage{colortbl}
\usepackage{tabularx}
\usepackage{algorithm}
\usepackage{algpseudocode}
\usepackage{placeins}
\usepackage{tcolorbox}
\tcbuselibrary{listings,breakable,skins}
\definecolor{promptpurple}{HTML}{5D597E}
\definecolor{promptbackground}{HTML}{F3F1F7}
\newtcblisting{promptbox}[1]{
    enhanced, listing only,
    colback=promptbackground, colframe=promptpurple,
    colbacktitle=promptbackground, coltitle=promptpurple,
    fonttitle=\bfseries\small, title={#1},
    boxrule=0.5pt, arc=2pt,
    left=5pt, right=5pt, top=4pt, bottom=4pt,
    listing options={basicstyle=\ttfamily\scriptsize,
        columns=fullflexible, keepspaces=true, breaklines=true, breakatwhitespace=true,
        showstringspaces=false, aboveskip=0pt, belowskip=0pt}
}

\definecolor{resultup}{HTML}{267A48}
\definecolor{resultdown}{HTML}{B44444}

\definecolor{cvprblue}{rgb}{0.21,0.49,0.74}
\usepackage[pagebackref,breaklinks,colorlinks,allcolors=cvprblue]{hyperref}

\def\confName{CVPR}
\def\confYear{2027}

\title{PathoJev: Efficient Pathology Decisions from Shared Visual Context}

\author{
{\normalsize Chengyang Zhang$^{1,2,\dagger}$, Wenchuan Zhang$^{2,\dagger}$, Bo Li$^{3}$, Xinyu Liu$^{1}$, Jiaming Yang$^{1}$, Mengran Li$^{4}$,}\\
{\normalsize Jie Chen$^{2}$, Wei Ju$^{1}$, Yuhao Yi$^{1,2,*}$, Hong Bu$^{2}$, Jiancheng Lv$^{1}$}\\[4pt]
{\small $^{1}$College of Computer Science, Sichuan University}\\
{\small $^{2}$Department of Pathology and Institute of Clinical Pathology, West China Hospital, Sichuan University}\\
{\small $^{3}$Department of Computer Science, School of Computing, National University of Singapore}\\
{\small $^{4}$School of Intelligent Systems Engineering, Sun Yat-sen University}\\[2pt]
{\small $^{\dagger}$Equal contribution.\quad $^{*}$Corresponding author: \texttt{yuhaoyi@scu.edu.cn}}
}

\hypersetup{pdftitle={PathoJev: Efficient Pathology Decisions from Shared Visual Context}}
\iftoggle{cvprfinal}{%
  \hypersetup{pdfauthor={Chengyang Zhang, Wenchuan Zhang, Bo Li, Xinyu Liu, Jiaming Yang, Mengran Li, Jie Chen, Wei Ju, Yuhao Yi, Hong Bu, Jiancheng Lv}}%
}{\hypersetup{pdfauthor={Anonymous}}}

\begin{document}
\maketitle
\begin{abstract}
Pathology workflows require multiple visual judgments about the same slide, from identifying tumor type to assessing grade and invasion.
Generating and parsing a separate textual response for each structured question adds latency between visual evidence and usable decisions.
Building on the System One paradigm, we introduce \textbf{PathoJev}, a pathology model that directly predicts answer probabilities for whole-slide and region-level assessment.
PathoJev encodes each slide once and evaluates independent question suffixes in batches over the shared visual context.
Reusing the hybrid backbone's recurrent states further limits the computation that grows with slide-context length for each additional question.
Across three whole-slide benchmarks, the 9B and 27B models achieve mean accuracies of \textbf{77.2\%} and \textbf{78.1\%}, exceeding SlideChat by \textbf{5.4} and \textbf{6.3} percentage points.
Both scales also maintain competitive accuracy on five ROI benchmarks.
On WSI-Bench, the 9B model runs \textbf{9.1$\boldsymbol{\times}$ faster} than SlideChat.
Shared context accelerates eight-question workloads by \textbf{6.9$\boldsymbol{\times}$} and \textbf{5.9$\boldsymbol{\times}$} over separate prefix encoding, rising to \textbf{11.2$\boldsymbol{\times}$} and \textbf{8.9$\boldsymbol{\times}$} for 64 questions.
These results provide a reusable probabilistic interface for combining multiple visual judgments in whole-slide and local tissue assessment.
Code: \url{https://github.com/ChyaZhang/PathoJev}.
\end{abstract}

\section{Introduction}
\label{sec:introduction}

Pathology assessment brings together multiple observations about tissue, including tumor type, histologic grade, and the presence of invasion. Synoptic reporting organizes specimen-level findings into explicit data elements and responses~\cite{cap2026protocols,schaad2024impact}. At the image level, several such judgments can draw on tissue represented in the same whole-slide image (WSI). Whole-slide visual question answering benchmarks capture this setting through questions about tissue morphology and diagnostic interpretation~\cite{chen2025slidechat,liang2025wsi}. This setting calls for reusing tissue evidence across questions that assess different aspects of the same slide.

\begin{figure*}[t!]
    \centering
    \includegraphics[width=\textwidth]{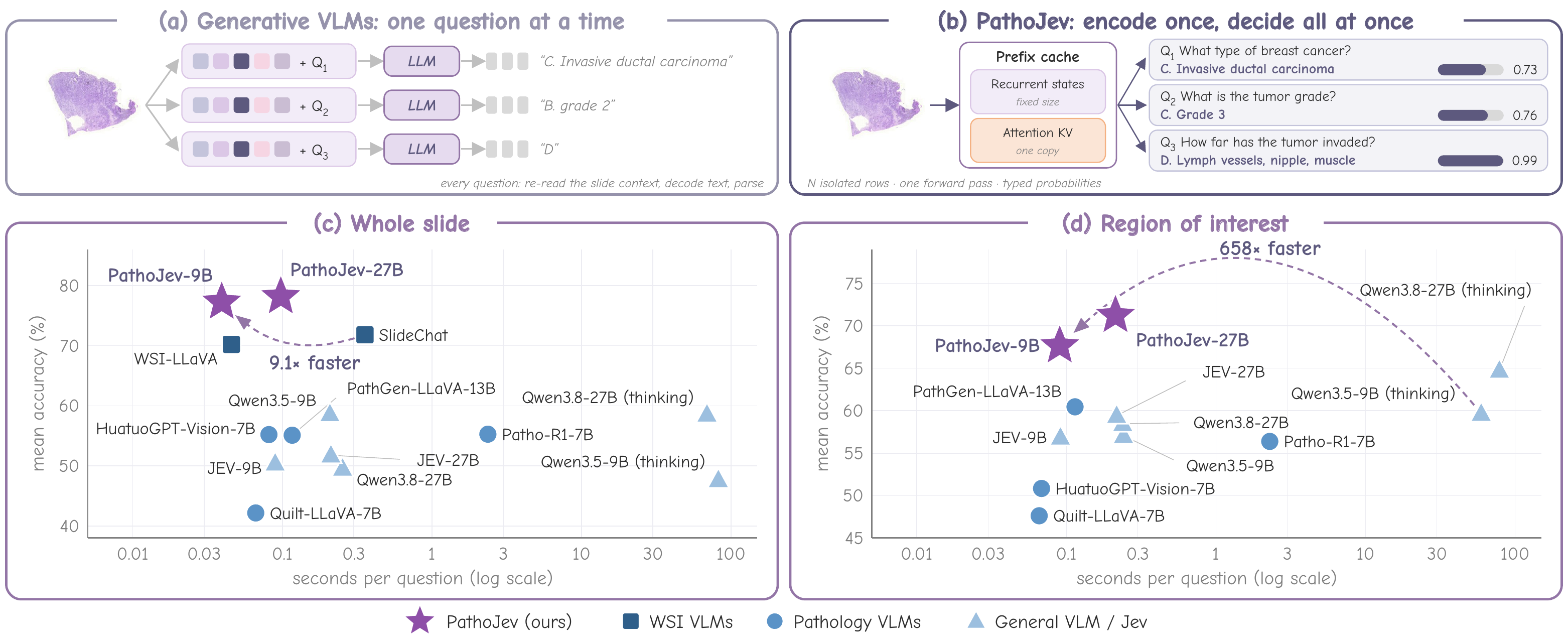}
    \caption{\textbf{PathoJev: shared visual context for multiple pathology decisions.}
    (a,b) Autoregressive answers versus direct probabilities from isolated questions sharing slide context. Re-reading context includes accessing cached prefix keys and values.
    (c,d) Whole-slide and ROI accuracy--latency comparisons. Accuracy averages each model's available results across three WSI or five ROI benchmarks. Latency is GPU forward time per question on WSI-Bench and PathMMU-tiny, respectively. Zero-shot WSI baselines use thumbnails.}
    \label{fig:overview}
\end{figure*}

Recent pathology vision-language models support instruction-following conversations about tissue images~\cite{seyfioglu2024quilt,sun2025pathgen,lu2024multimodal}, whole-slide question answering and report generation~\cite{guo2024histgen,chen2025slidechat,liang2025wsi}, and explicit diagnostic reasoning~\cite{zhang2026patho}. These models express their answers through autoregressive text generation. For questions with specified alternatives, obtaining a decision therefore requires sequential decoding followed by answer extraction (Figure~\ref{fig:overview}a). The generated text also does not directly expose a normalized distribution over the alternatives. When several questions concern the same slide, this generation process is repeated for each judgment, even though the underlying tissue evidence is shared.

The shared evidence and compact answer spaces suggest separating visual context processing from question-specific decisions. The System One paradigm provides this formulation by mapping an input state directly to typed probabilistic judgments~\cite{almeida2026introducing}, with recent extensions to shared image and video context~\cite{yu2026visual,zhu2026videojev}. Whole-slide pathology makes the cost of accessing that context particularly relevant: tissue evidence spans a long sequence of patch representations, while each question may require only a binary or multiple-choice answer. Caching the prefix saves repeated prefill, but softmax attention still reads its keys and values for each new question. Efficient whole-slide decisions thus depend on both removing answer decoding and controlling the cost of reusing a long visual context.

Building on the System One paradigm, we introduce PathoJev, a pathology model for multiple decisions over shared tissue evidence (Figure~\ref{fig:overview}b). A slide input pathway aggregates spatial patch features into a shared visual prefix. Questions branch independently from this prefix and are evaluated in batches, with a typed head returning answer probabilities directly. Reusing the hybrid backbone's recurrent states reduces the portion of question processing that grows with slide-context length. PathoJev also retains a region-of-interest (ROI) input pathway and learns from both whole-slide and ROI supervision, bringing slide-level and local tissue judgments into the same decision framework.

We evaluate PathoJev at 9B and 27B scales on three whole-slide and five ROI benchmarks. The 9B and 27B models improve mean whole-slide accuracy over SlideChat by 5.4 and 6.3 percentage points, and over WSI-LLaVA by 7.0 and 7.9 points, respectively. PathoJev-9B is $9.1\times$ faster than SlideChat on WSI-Bench. A same-data autoregressive control further tests the decision interface: PathoJev-9B achieves higher accuracy on six of eight benchmarks, with $3.0\times$ lower WSI latency and $2.4\times$ lower ROI latency. For eight-question workloads, shared context reduces total time by $6.9\times$ at 9B and $5.9\times$ at 27B relative to separate prefix encoding, rising to $11.2\times$ and $8.9\times$ for 64 questions. These comparisons connect pathology performance to the complementary benefits of direct readout and context reuse.

Our contributions are as follows:
\begin{itemize}
    \item We introduce PathoJev, a System One model that supports whole-slide and ROI assessment through a unified interface for language-conditioned pathology questions and direct probabilistic decisions.
    \item We combine spatial patch aggregation with independent question branches that reuse attention caches and recurrent states, and analyze the cost of additional decisions over long slide contexts.
    \item We evaluate 9B and 27B models on three whole-slide and five ROI benchmarks, showing accuracy gains and latency reductions against a range of general-purpose and pathology-specific baselines, and analyze inference cost as question count and slide-context length increase.
\end{itemize}

\section{Related Work}
\label{sec:related_work}

\paragraph{Pathology Vision-Language Models.}
Pathology vision-language pretraining connects tissue morphology with textual descriptions, enabling recognition and retrieval through a shared semantic space~\cite{huang2023visual,lu2024visual}. Instruction tuning extends these representations to pathology conversations and visual question answering. For example, Quilt-LLaVA constructs visual instruction data from localized narratives in educational videos~\cite{seyfioglu2024quilt}, while PathChat adapts a pretrained vision-language model for pathology conversations~\cite{lu2024multimodal}. Synthetic image-text pairs provide another source of supervision for pathology instruction tuning~\cite{sun2025pathgen}. Patho-R1 further develops diagnostic reasoning through chain-of-thought supervision and reinforcement learning~\cite{zhang2026patho}. These conversational models typically express their predictions through autoregressive generation, including when evaluation requires selecting from predefined answers. PathoJev retains language-defined pathology questions and predicts probabilities over their answer alternatives directly.

\paragraph{Whole-Slide Image Understanding.}
Whole-slide analysis requires aggregating information from large collections of tissue patches. Multiple instance learning uses slide-level supervision for this purpose: CLAM identifies diagnostically relevant regions through attention-based aggregation~\cite{lu2021data}, while correlation-based approaches model relationships among patches~\cite{shao2021transmil}. Multimodal whole-slide pretraining further aligns slide representations with captions and pathology reports, supporting transfer to classification and retrieval~\cite{ding2025multimodal}. Language-driven approaches extend this setting to report generation through local-global feature aggregation and cross-modal interaction~\cite{guo2024histgen}. SlideChat and WSI-LLaVA combine slide-level visual encoding with instruction tuning to support whole-slide question answering~\cite{chen2025slidechat,liang2025wsi}. Building on the ability to represent slide-level evidence, PathoJev addresses the repeated computation needed to evaluate multiple questions over that evidence.

\paragraph{System One Models.}
System One models expose typed probabilistic judgments as outputs that software can consume directly. Jev introduces this formulation for decisions over specified output spaces, replacing answer-string generation with structured predictions~\cite{almeida2026introducing}. Visual Jev extends direct decision making to multiple questions about one image, sharing visual context across isolated question suffixes and reading candidate probabilities from the language-model head~\cite{yu2026visual}. VideoJev uses a shared scoring head for temporal intervals and textual answer options, with cached and joint inference supporting repeated queries over video evidence~\cite{zhu2026videojev}. PathoJev adapts shared-context decision inference to spatially aggregated whole-slide features and learns a common decision interface from WSI and ROI supervision. Its computational analysis relates the cost of additional pathology questions to prefix reuse and the hybrid backbone's recurrent states, examining both question count and slide-context length.

\section{Method}
\label{sec:method}

\begin{figure*}[t]
    \centering
    \includegraphics[width=\textwidth]{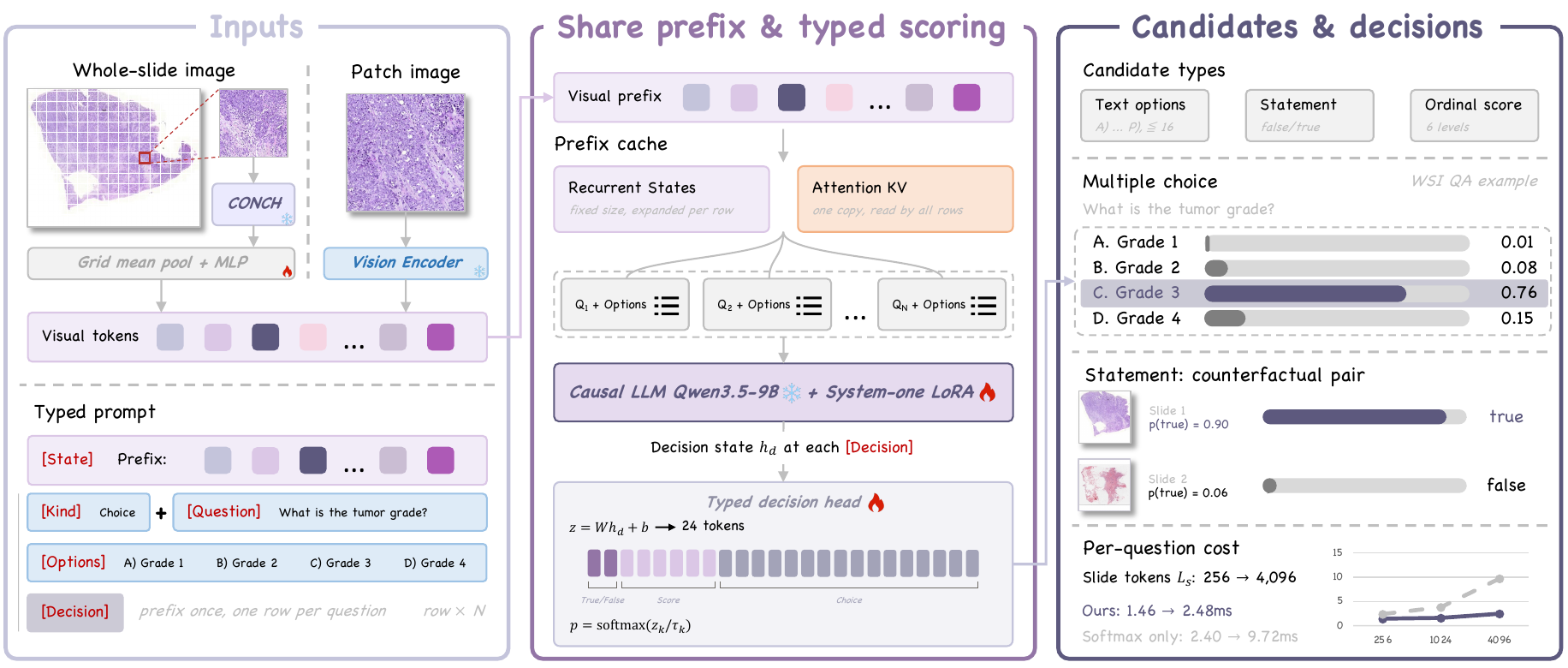}
    \caption{\textbf{Overview of PathoJev.} Left: WSI patch features and ROI images enter through separate visual pathways. Center: WSI question rows reuse a shared visual prefix independently, with fixed-size recurrent states and shared attention keys and values; a typed head returns answer probabilities. Right: supported output types, test-set decision examples, and marginal computation per additional question as slide-context length increases. The marginal-cost comparison uses Qwen3-8B as a softmax-only baseline.}
    \label{fig:architecture}
\end{figure*}

PathoJev combines direct probability readout with reusable slide representations for efficient decisions over long visual contexts (Figure~\ref{fig:architecture}). We define a common WSI and ROI decision interface, describe independent question processing over a shared prefix, and analyze its computational cost.

\subsection{A Unified Pathology Decision Framework}
\label{sec:decision_framework}

\paragraph{Problem formulation.}
Given a pathology image $X$, we consider $N$ questions $\{q_i\}_{i=1}^{N}$. Each question specifies an output type $t_i$ and an ordered answer set $\mathcal{O}_i=\{o_{i1},\ldots,o_{iK_i}\}$. The image can be a WSI or an ROI. The model predicts
\begin{equation}
    p_{ik}=p_{\theta}(o_{ik}\mid X,q_i,\mathcal{O}_i,t_i),
    \qquad k=1,\ldots,K_i,
    \label{eq:decision_distribution}
\end{equation}
and selects the answer with the highest probability. Multiple-choice questions provide textual alternatives, while binary judgments use a two-element answer set. For example, questions about the same image may ask for a histologic grade among named categories and whether invasion is visible. The question and its alternatives specify the judgment, while the image supplies the tissue evidence. Equation~\ref{eq:decision_distribution} makes each prediction conditional on that evidence and its own question, allowing a set of independent judgments to describe different aspects of the same tissue.

\paragraph{Visual inputs.}
For a WSI, we extract features from $512\times512$ tissue patches at $20\times$ magnification using the frozen CONCH v1.5 encoder~\cite{ding2025multimodal}. Spatial mean pooling on a $32\times32$ grid aggregates the patch features, and a trainable multilayer perceptron projects the pooled features into the language backbone's embedding space:
\begin{equation}
    \begin{alignedat}{1}
        \bar f_c &= \frac{1}{|\mathcal{I}_c|}\sum_{j\in\mathcal{I}_c}f_{\mathrm{CONCH}}(x_j),\\
        V_X &= g_{\phi}\!\left((\bar f_c)_{c\in\mathcal{G}_X}\right),
    \end{alignedat}
    \label{eq:slide_encoding}
\end{equation}
where $\mathcal{G}_X$ is the ordered set of occupied grid cells and $\mathcal{I}_c$ indexes patches located in cell $c$. Grid pooling preserves a spatially organized representation while reducing the visual sequence length. For ROI inputs, we retain the backbone's native visual pathway. ROI questions address local tissue findings through the same typed decision interface used for slide-level assessment. Both pathways produce visual tokens for decision making, and their visual encoders remain frozen during training.

\paragraph{Structured question interface.}
Following the System One formulation~\cite{almeida2026introducing}, we express each judgment using ordinary text markers: \texttt{[state]} introduces the image context, \texttt{[kind]} specifies the output type, \texttt{[question]} introduces the question, and \texttt{[options]} introduces its ordered alternatives. Each question ends with \texttt{[decision]:}. For WSI inputs, the state precedes all question-specific fields, enabling a shared prefix:
\begin{equation}
    \begin{alignedat}{1}
        P_X &= E(c_{\mathrm{WSI}})\,\Vert\,V_X\,\Vert\,E(\texttt{\textbackslash n}),\\
        R_i &= \texttt{[kind]}\;t_i\;\texttt{[question]}\;q_i\\
            &\quad\texttt{[options]}\;\mathcal{O}_i\;\texttt{[decision]:}.
    \end{alignedat}
    \label{eq:typed_prompt}
\end{equation}
Here, $E$ embeds text and $c_{\mathrm{WSI}}$ denotes the WSI state prompt. ROI inputs place \texttt{[kind]} before \texttt{[state]} and process one question per image. Full prompts are provided in Appendix~\ref{app:prompt_format}.

\paragraph{Typed decision readout.}
Let $h_i\in\mathbb{R}^{d}$ denote the hidden state after the backbone's final normalization at the last token of \texttt{[decision]:}. A shared linear head produces $z_i=W_dh_i+b_d\in\mathbb{R}^{24}$. We represent the output space by 24 typed decision tokens, each associated with one logit of this head: 16 for textual choices, two for binary judgments, and six for ordinal scores. The output type selects the corresponding token group, and the answer set determines which tokens are valid for the current question. Writing $a_t(k)$ for the output index associated with answer $k$ of type $t$, we normalize over its $K_i$ valid candidates:
\begin{equation}
    p_{ik}=\frac{\exp\!\left(z_{i,a_{t_i}(k)}/\tau_{t_i}\right)}
    {\sum_{j=1}^{K_i}\exp\!\left(z_{i,a_{t_i}(j)}/\tau_{t_i}\right)}.
    \label{eq:typed_readout}
\end{equation}
Here, $\tau_{t_i}>0$ denotes the temperature for output type $t_i$. The text in \texttt{[options]} supplies the meaning of each choice token for the current question. Our pathology tasks use multiple-choice and binary decisions. A single readout at this final position returns their answer probabilities directly, supporting confidence-based selection without answer decoding.

\subsection{Shared Slide Context with Isolated Questions}
\label{sec:shared_context}

Several questions may depend on the same tissue evidence. PathoJev therefore separates computation of the visual prefix from computation of the question suffixes. For the layout in Eq.~\ref{eq:typed_prompt}, we prefill the shared state prefix $P_X$ once to obtain its layer-wise cache $\mathcal{C}_X$. Each question suffix $R_i$ then branches from this same cache:
\begin{equation}
    \begin{alignedat}{1}
        \mathcal{C}_X &= \operatorname{Prefill}_{\theta}(P_X), \\
        h_i &= \operatorname{Read}_{\theta}(R_i;\mathcal{C}_X).
    \end{alignedat}
    \label{eq:shared_read}
\end{equation}
Here, $\operatorname{Read}_{\theta}$ returns the hidden state at the suffix's decision position. Following shared-context visual decision inference~\cite{yu2026visual}, questions are processed as independent rows. A token in row $i$ can attend to the shared prefix and preceding tokens within that row. The prefix is computed before the questions, and each suffix reads only its own local history. Rows can therefore be evaluated in batches while preserving the conditioning structure in Eq.~\ref{eq:decision_distribution}. Adding or reordering questions leaves each row's visual and textual context unchanged.

The hybrid backbone combines recurrent Gated DeltaNet layers~\cite{yang2025gated} with softmax-attention layers. Each recurrent layer summarizes the prefix in fixed-size recurrent and convolutional states. These states initialize separate question branches, whose updates remain local to each row. Each softmax-attention layer instead retains a shared set of prefix keys and values, which a row reads together with its own suffix keys and values. The prefix cache remains unchanged throughout question evaluation. This organization reuses the slide computation while maintaining independent question contexts during both training and inference. The different cache structures also determine how the cost of an additional question depends on slide length.

\subsection{Efficient Multi-Question Inference}
\label{sec:inference_cost}

Sharing the prefix amortizes slide processing across questions about the same image. For a prefix of length $L$ and question suffix lengths $\{m_i\}_{i=1}^{N}$, the computational work can be decomposed as
\begin{equation}
    \begin{aligned}
    C_{\mathrm{total}}={}&C_{\mathrm{visual}}(X)+C_{\mathrm{prefill}}(L)\\
    &+\sum_{i=1}^{N}C_{\mathrm{row}}(L,m_i),
    \end{aligned}
    \label{eq:total_cost}
\end{equation}
where $C_{\mathrm{visual}}$ and $C_{\mathrm{prefill}}$ denote visual encoding and prefix prefill, respectively. Both are incurred once per image. For a fixed slide, their contribution to average per-question cost decreases as the number of questions increases. Each row requires a single forward evaluation of its suffix and decision head, with no additional answer-decoding steps. Batched execution processes these independent rows together.

The hybrid backbone also reduces the portion of each row's computation that depends on prefix length. Recurrent layers process the suffix from a state whose size is independent of $L$. Softmax-attention layers still read the shared prefix keys and values. For fixed model dimensions, the per-row cost takes the form
\begin{equation}
    C_{\mathrm{row}}(L,m)=C_{\mathrm{local}}(m)
    +\mathcal{O}(n_{\mathrm{att}}mLd_{\mathrm{att}}),
    \label{eq:row_cost}
\end{equation}
where $n_{\mathrm{att}}$ is the number of softmax-attention layers and $d_{\mathrm{att}}$ is their effective attention width. The term $C_{\mathrm{local}}$ includes recurrent updates, suffix-local attention, projections, and feed-forward computation, all independent of $L$. The 9B model contains 24 recurrent and eight softmax-attention layers, while the 27B model contains 48 and 16, respectively~\cite{qwen2026qwen35,qwen2026qwen38}. Thus, prefix-length-dependent attention is restricted to one quarter of the layers. Shared prefill and direct readout address repeated computation across questions, while recurrent state reuse limits where slide length affects the cost of each additional question.

\subsection{Training Objective and Two-Stage Adaptation}
\label{sec:training_recipe}

\paragraph{Decision supervision.}
Both adaptation stages supervise the probability assigned to the correct answer. For a question with target index $y_i$, the cross-entropy over its valid candidates is
\begin{equation}
    \ell_i=-\sum_{k=1}^{K_i}\mathbf{1}[k=y_i]\log p_{ik}
    =-\log p_{i,y_i},
    \label{eq:training_loss}
\end{equation}
where $p_{ik}$ is defined by Eq.~\ref{eq:typed_readout}. A binary judgment is the $K_i=2$ case of the same objective. Questions sharing an image receive separate supervision through their own decision positions. For a training distribution $\mathcal{D}$, we write the expected question loss as
\begin{equation}
    \mathcal{L}_{\mathcal{D}}(\phi,\Delta,\psi)
    =\mathbb{E}_{i\sim\mathcal{D}}[\ell_i(\phi,\Delta,\psi)],
    \label{eq:training_risk}
\end{equation}
where $\phi$ denotes the WSI projection parameters, $\Delta$ the LoRA parameters~\cite{hu2022lora}, and $\psi=(W_d,b_d)$ the decision-head parameters. The distribution $\mathcal{D}$ reflects question sampling during training.

\paragraph{Stage A: slide projection alignment.}
We initialize the decision model from pretrained JEV checkpoints~\cite{autotrust2026jev9b,autotrust2026jev27b}, with LoRA parameters $\Delta_0$ and head parameters $\psi_0$. The first stage learns the WSI projection from slide questions while keeping the decision model fixed:
\begin{equation}
    \min_{\phi}\;\mathcal{L}_{\mathcal{D}_{A}}(\phi,\Delta_0,\psi_0),
    \label{eq:stage_a}
\end{equation}
where $\mathcal{D}_{A}$ samples WSI training questions. Gradients from the correct-answer probability align pooled tissue features with the pretrained decision representation. This establishes the slide input pathway before adapting the question-processing components.

\paragraph{Stage B: joint pathology adaptation.}
Starting from the aligned projection, we jointly optimize the WSI projection, LoRA adapters, and typed head:
\begin{equation}
    \min_{\phi,\Delta,\psi}\;\mathcal{L}_{\mathcal{D}_{B}}(\phi,\Delta,\psi),
    \label{eq:stage_b}
\end{equation}
where $\mathcal{D}_{B}$ mixes WSI and ROI training questions. WSI supervision updates the slide projection and shared decision components, while ROI supervision adapts the same LoRA parameters and head through the native visual pathway. The pretrained language-backbone weights and visual encoders remain frozen in both stages. Joint adaptation therefore connects slide-level and local tissue judgments through a common probabilistic interface. Sampling schedules and optimization hyperparameters are given in Appendix~\ref{app:experimental_details}.

\section{Experiments}
\label{sec:experiments}

\begin{table*}[t]
    \centering
    \caption{Benchmark accuracy (\%, $\uparrow$) and inference latency (s/q, $\downarrow$). Avg. is the unweighted mean over five ROI or three whole-slide benchmarks. Best and second-best results within each model-size group are \textbf{bold} and \underline{underlined}, with ties marked equally. Latency is measured on PathMMU-tiny for ROI inputs and WSI-Bench for whole-slide inputs, at batch size one.}
    \label{tab:main_results}
    \begingroup
    \footnotesize
    \setlength{\tabcolsep}{2.3pt}
    \renewcommand{\arraystretch}{1.18}
    \begin{tabularx}{\textwidth}{>{\raggedright\arraybackslash}X*{12}{c}}
        \toprule
        \multirow{3}{*}{\textbf{Model}} & \multicolumn{7}{c}{\textbf{Region of Interest}} & \multicolumn{5}{c}{\textbf{Whole Slide}} \\
        \cmidrule(lr){2-8}\cmidrule(lr){9-13}
        & \makecell[c]{PathMMU\\tiny} & \makecell[c]{PathVQA} & \makecell[c]{Quilt\\VQA} & \makecell[c]{OmniMed\\VQA} & \makecell[c]{MedXpert\\QA} & \makecell[c]{\textbf{Avg.}} & \makecell[c]{Latency\\(s/q)} & \makecell[c]{WSI\\VQA} & \makecell[c]{WSI\\Bench} & \makecell[c]{SlideBench\\TCGA} & \makecell[c]{\textbf{Avg.}} & \makecell[c]{Latency\\(s/q)} \\
        \midrule
        \multicolumn{13}{l}{\textcolor{black!65}{\textbf{7--13B models}}} \\
        Patho-R1-7B (reasoning)~\cite{zhang2026patho}$^{\dagger}$ & 67.6 & 57.1 & \textbf{70.8} & \underline{69.7} & 16.7 & 56.4 & 2.291 & 44.1 & 67.8 & 53.9 & 55.3 & 2.367 \\
        HuatuoGPT-Vision-7B~\cite{chen2024injecting}$^{\dagger}$ & 59.5 & 66.6 & 64.4 & 51.5 & 12.2 & 50.8 & \underline{0.068} & 47.7 & 66.8 & 51.1 & 55.2 & 0.081 \\
        Quilt-LLaVA-v1.5-7B~\cite{seyfioglu2024quilt}$^{\dagger}$ & 46.4 & 56.7 & \underline{70.3} & 46.9 & 17.8 & 47.6 & \textbf{0.066} & 39.7 & 49.4 & 37.4 & 42.2 & 0.066 \\
        PathGen-LLaVA-13B~\cite{sun2025pathgen}$^{\dagger}$ & 71.4 & 62.7 & 68.2 & 68.9 & \textbf{31.1} & \underline{60.5} & 0.114 & 43.8 & 69.1 & 52.4 & 55.1 & 0.116 \\
        SlideChat~\cite{chen2025slidechat} & \textemdash & \textemdash & \textemdash & \textemdash & \textemdash & \textemdash & \textemdash & \underline{59.7} & 81.8 & \underline{73.9} & \underline{71.8} & 0.357 \\
        WSI-LLaVA~\cite{liang2025wsi} & \textemdash & \textemdash & \textemdash & \textemdash & \textemdash & \textemdash & \textemdash & 50.5 & \underline{90.0} & 70.1 & 70.2 & \underline{0.046} \\
        Qwen3.5-9B (generate)$^{\dagger}$ & 69.9 & \underline{77.8} & 65.0 & 62.6 & 10.0 & 57.1 & 0.241 & 48.5 & 74.4 & 53.0 & 58.6 & 0.208 \\
        Qwen3.5-9B (thinking)$^{\dagger}$ & 66.7 & 74.4 & 70.0 & 63.9 & 23.3 & 59.7 & 59.4 & 40.3 & 59.2 & 43.5 & 47.7 & 82.3 \\
        JEV-9B (zero-shot)$^{\dagger}$ & \textbf{72.3} & 72.4 & 57.7 & 60.8 & 21.1 & 56.9 & 0.091 & 38.7 & 65.2 & 47.4 & 50.4 & 0.089 \\
        \rowcolor{blue!10}
        \textbf{PathoJev-9B} & \underline{71.9} & \textbf{94.6} & \textbf{70.8} & \textbf{75.3} & \underline{25.6} & \textbf{67.6} & 0.090 & \textbf{65.4} & \textbf{90.4} & \textbf{75.9} & \textbf{77.2} & \textbf{0.039} \\
        \midrule
        \multicolumn{13}{l}{\textcolor{black!65}{\textbf{27B models}}} \\
        Qwen3.8-27B (generate)$^{\dagger}$ & 64.5 & 69.2 & \textbf{68.2} & 46.0 & \underline{44.4} & 58.5 & 0.238 & 41.3 & 59.7 & 47.7 & 49.6 & 0.253 \\
        Qwen3.8-27B (thinking)$^{\dagger}$ & 70.3 & \underline{70.5} & 62.4 & \textbf{70.5} & \textbf{50.0} & \underline{64.7} & 78.6 & \underline{48.7} & \underline{71.3} & \underline{55.7} & \underline{58.6} & 69.1 \\
        JEV-27B-VL (zero-shot)$^{\dagger}$ & \textbf{74.5} & 68.3 & 55.7 & 59.7 & 38.9 & 59.4 & \underline{0.217} & 37.9 & 67.4 & 50.0 & 51.8 & \underline{0.211} \\
        \rowcolor{blue!10}
        \textbf{PathoJev-27B} & \underline{73.8} & \textbf{95.1} & \underline{67.3} & \underline{69.9} & \textbf{50.0} & \textbf{71.2} & \textbf{0.213} & \textbf{66.4} & \textbf{91.1} & \textbf{76.9} & \textbf{78.1} & \textbf{0.098} \\
        \bottomrule
    \end{tabularx}
    \par\smallskip
    \begin{minipage}{\textwidth}
        \footnotesize
        $\dagger$ WSI evaluation uses slide thumbnails. Thinking latency uses the first 100 questions in test-file order.
    \end{minipage}
    \endgroup
\end{table*}

\subsection{Experimental Setup}
\label{sec:experimental_setup}

\smallskip\noindent\textbf{Implementation details.}
We train PathoJev at 9B and 27B scales using the two-stage schedule in Sec.~\ref{sec:training_recipe}, with AdamW and rank-16 LoRA adapters on four RTX PRO 6000 GPUs with 96\,GB memory each. Slide projection alignment runs for one epoch, followed by two epochs of joint pathology adaptation. In the second stage, learning rates are $2\times10^{-4}$ for the WSI projection and $10^{-4}$ for the LoRA adapters and decision head. Whole-slide supervision combines SlideInstruct~\cite{chen2025slidechat}, WSI-Bench~\cite{liang2025wsi}, WSI-VQA~\cite{chen2025wsivqa}, and CF-Path, a pathology training set we construct from TCGA reports to provide slide-level binary statement supervision. ROI supervision uses PathVQA~\cite{he2021towards} and the pathology subset of PMC-VQA~\cite{zhang2024development}. Data composition and detailed training settings are provided in Appendix~\ref{app:experimental_details}.

\smallskip\noindent\textbf{Evaluation.}
We evaluate whole-slide decisions on WSI-VQA~\cite{chen2025wsivqa}, closed-ended WSI-Bench~\cite{liang2025wsi}, and SlideBench-TCGA~\cite{chen2025slidechat}. Local tissue assessment uses PathMMU-tiny~\cite{sun2024pathmmu}, the yes/no subset of PathVQA~\cite{he2021towards}, closed-ended Quilt-VQA~\cite{seyfioglu2024quilt}, and the pathology subsets of OmniMedVQA~\cite{hu2024omnimedvqa} and MedXpertQA~\cite{zuo2025medxpertqa}. Evaluation subsets, sample counts, and baseline models are detailed in Appendix~\ref{app:evaluation}.

\smallskip\noindent\textbf{Metrics.}
We report question-level accuracy and compute unweighted benchmark means separately for WSI and ROI evaluation. PathoJev selects the highest-probability answer; generative outputs are parsed into the supplied alternatives, with unparseable responses counted as incorrect. Probability quality is assessed using expected calibration error (ECE)~\cite{guo2017calibration} with 15 bins and top-1 confidence. Confidence-based prediction retention is evaluated using coverage--accuracy curves and the area under the risk--coverage curve (AURC), as detailed in Appendix~\ref{app:probability_metrics}.

\smallskip\noindent\textbf{Inference latency.}
We measure synchronized forward time after untimed warm-up on an exclusively allocated RTX PRO 6000 GPU with 96\,GB memory. Main comparisons use batch size one and report seconds per question on WSI-Bench and PathMMU-tiny. For feature-based WSI models, timing excludes offline visual feature extraction. ROI timing includes visual encoding but excludes image decoding and preprocessing. The operations timed for each WSI model are specified in Appendix~\ref{app:latency_protocol}. PathoJev, SlideChat, and WSI-LLaVA reuse the slide prefix.

\begin{table*}[t]
    \centering
    \caption{Decision readout versus autoregressive fine-tuning at 9B. Both models use the Qwen3.5-9B backbone, the same training data, two-stage schedule, and step counts; PathoJev starts from JEV-9B and the autoregressive model from Qwen3.5-9B. Accuracy is reported in percent and latency in seconds per question. Avg. follows Table~\ref{tab:main_results}. Best and second-best results are \textbf{bold} and \underline{underlined}.}
    \label{tab:readout_ablation}
    \begingroup
    \footnotesize
    \setlength{\tabcolsep}{2.3pt}
    \renewcommand{\arraystretch}{1.18}
    \begin{tabularx}{\textwidth}{>{\raggedright\arraybackslash}X*{12}{c}}
        \toprule
        \multirow{3}{*}{\textbf{Variant}} & \multicolumn{7}{c}{\textbf{Region of Interest}} & \multicolumn{5}{c}{\textbf{Whole Slide}} \\
        \cmidrule(lr){2-8}\cmidrule(lr){9-13}
        & \makecell[c]{PathMMU\\tiny} & \makecell[c]{PathVQA} & \makecell[c]{Quilt\\VQA} & \makecell[c]{OmniMed\\VQA} & \makecell[c]{MedXpert\\QA} & \makecell[c]{\textbf{Avg.}} & \makecell[c]{Latency\\(s/q)} & \makecell[c]{WSI\\VQA} & \makecell[c]{WSI\\Bench} & \makecell[c]{SlideBench\\TCGA} & \makecell[c]{\textbf{Avg.}} & \makecell[c]{Latency\\(s/q)} \\
        \midrule
        Autoregressive SFT & \textbf{74.3} & \underline{93.8} & \underline{66.8} & \underline{68.8} & \textbf{30.0} & \underline{66.7} & \underline{0.215} & \underline{63.8} & \underline{89.0} & \underline{74.2} & \underline{75.7} & \underline{0.119} \\
        \rowcolor{blue!10}
        \textbf{PathoJev-9B} & \underline{71.9} & \textbf{94.6} & \textbf{70.8} & \textbf{75.3} & \underline{25.6} & \textbf{67.6} & \textbf{0.090} & \textbf{65.4} & \textbf{90.4} & \textbf{75.9} & \textbf{77.2} & \textbf{0.039} \\
        \midrule
        $\Delta$ / speedup & \textcolor{resultdown}{$-2.4$} & \textcolor{resultup}{$+0.8$} & \textcolor{resultup}{$+4.1$} & \textcolor{resultup}{$+6.5$} & \textcolor{resultdown}{$-4.4$} & \textcolor{resultup}{$+0.9$} & \textcolor{resultup}{$2.4\times$} & \textcolor{resultup}{$+1.5$} & \textcolor{resultup}{$+1.4$} & \textcolor{resultup}{$+1.7$} & \textcolor{resultup}{$+1.5$} & \textcolor{resultup}{$3.0\times$} \\
        \bottomrule
    \end{tabularx}
    \par\smallskip
    \begin{minipage}{\textwidth}
        \footnotesize
        $\Delta$ denotes PathoJev minus autoregressive SFT in percentage points; latency columns give speedups. Paired-bootstrap 95\% confidence intervals are reported in Appendix~\ref{app:readout_ablation}.
    \end{minipage}
    \endgroup
\end{table*}

\subsection{Main Results}
\label{sec:main_results}

Table~\ref{tab:main_results} compares PathoJev with general-purpose and pathology-specific models on three whole-slide and five ROI benchmarks. We examine whether direct probabilistic decisions improve the accuracy--latency trade-off at both scales, from repeated judgments over shared slide context to local tissue assessment.

\smallskip\noindent\textbf{Whole-slide assessment.}
PathoJev achieves the highest accuracy among evaluated models in each size group on all three WSI benchmarks. Mean accuracy reaches 77.2\% at 9B and 78.1\% at 27B, compared with 71.8\% for SlideChat and 70.2\% for WSI-LLaVA. Relative to their respective Qwen backbones in generation mode, the two scales improve mean accuracy by 18.6 and 28.5 percentage points. PathoJev-9B also has the lowest measured WSI latency at 0.039 seconds per question, yielding a $9.1\times$ speedup over SlideChat. Since the generative WSI baselines also reuse the slide-prefix cache, this advantage extends beyond avoiding repeated visual prefill. Direct readout produces each decision without an autoregressive answer sequence, supporting efficient repeated judgments over the same tissue context.

\smallskip\noindent\textbf{ROI assessment.}
PathoJev also improves local tissue assessment, reaching mean accuracies of 67.6\% and 71.2\% at 9B and 27B, respectively. These correspond to gains of 10.5 and 12.7 points over their Qwen backbones in generation mode, with $2.7\times$ and $1.1\times$ speedups on PathMMU-tiny. Within the 7--13B group, PathoJev-9B leads on three of five benchmarks, including a tie on Quilt-VQA. Gains vary by task, with more consistent scaling improvements in whole-slide assessment. The same probabilistic interface supports both slide-level and local tissue judgments, illustrated in Appendix~\ref{app:qualitative}.

\smallskip\noindent\textbf{Comparison with explicit reasoning.}
Relative to their Thinking-mode backbones, PathoJev-9B and PathoJev-27B improve mean ROI accuracy by 7.9 and 6.5 points, with approximately $658\times$ and $369\times$ speedups on PathMMU-tiny, respectively. At 27B, mean WSI accuracy also increases from 58.6\% to 78.1\%, with an approximately $705\times$ speedup on WSI-Bench. This comparison highlights the efficiency of matching the output interface to the task: for these structured judgments, PathoJev maps tissue evidence directly to answer probabilities without generating a reasoning trace or parsing a textual answer.

\subsection{Ablation Study}
\label{sec:ablations}

We examine the decision interface and reuse of visual context through two comparisons. The first matches the 9B backbone, pathology supervision, and training steps; the second varies prefix reuse while holding the trained model and questions fixed at both scales.

\noindent\textbf{Decision readout versus autoregressive fine-tuning.}
Table~\ref{tab:readout_ablation} compares PathoJev-9B with an autoregressive model using the same backbone, training data, and number of training steps. The control learns textual answers and uses greedy decoding with a 16-token output limit. PathoJev reduces latency by $3.0\times$ on WSI-Bench and $2.4\times$ on PathMMU-tiny while retaining competitive accuracy, with higher point estimates on six of eight benchmarks. Mean accuracy increases by 1.5 points on WSI tasks and 0.9 points on ROI tasks. Direct readout therefore combines efficient decisions with normalized answer probabilities. Appendix~\ref{app:readout_ablation} reports per-benchmark differences and paired confidence intervals.

\begin{table}[t]
    \centering
    \caption{Prefix-sharing ablation. Median time (seconds per slide, $\downarrow$) over 40 WSI-Bench slides with $N$ questions. Speedups compare shared with separate encoding at $N=64$. Questions are repeated when $N$ exceeds the available set.}
    \label{tab:prefix_ablation}
    \begingroup
    \footnotesize
    \setlength{\tabcolsep}{3pt}
    \renewcommand{\arraystretch}{1.18}
    \begin{tabularx}{\columnwidth}{>{\raggedright\arraybackslash}Xcccc}
        \toprule
        \textbf{Setting} & $N=1$ & $N=8$ & $N=64$ & \textbf{Speedup} \\
        \midrule
        \multicolumn{5}{l}{\textcolor{black!65}{\textbf{PathoJev-9B}}} \\
        Separate encoding & \textbf{0.12} & \underline{0.95} & \underline{7.63} & \textemdash \\
        \rowcolor{blue!10}
        Shared prefix & \textbf{0.12} & \textbf{0.14} & \textbf{0.68} & $11.2\times$ \\
        \midrule
        \multicolumn{5}{l}{\textcolor{black!65}{\textbf{PathoJev-27B}}} \\
        Separate encoding & \textbf{0.28} & \underline{2.23} & \underline{17.86} & \textemdash \\
        \rowcolor{blue!10}
        Shared prefix & \textbf{0.28} & \textbf{0.38} & \textbf{2.01} & $8.9\times$ \\
        \bottomrule
    \end{tabularx}
    \endgroup
\end{table}

\smallskip\noindent\textbf{Shared slide context.}
We compare shared-prefix inference with separately encoding the slide for each question (Table~\ref{tab:prefix_ablation}). For 9B, increasing the workload from one to eight questions changes shared-prefix time from 0.12 to 0.14 seconds, whereas separate encoding reaches 0.95 seconds. At 64 questions, sharing reduces total time from 7.63 to 0.68 seconds at 9B and from 17.86 to 2.01 seconds at 27B. These $11.2\times$ and $8.9\times$ reductions show how reusing slide context amortizes prefix processing across decisions. Appendix~\ref{app:efficiency_analysis} details the repeated-question workloads.

\begin{figure*}[t]
    \centering
    \includegraphics[width=\textwidth]{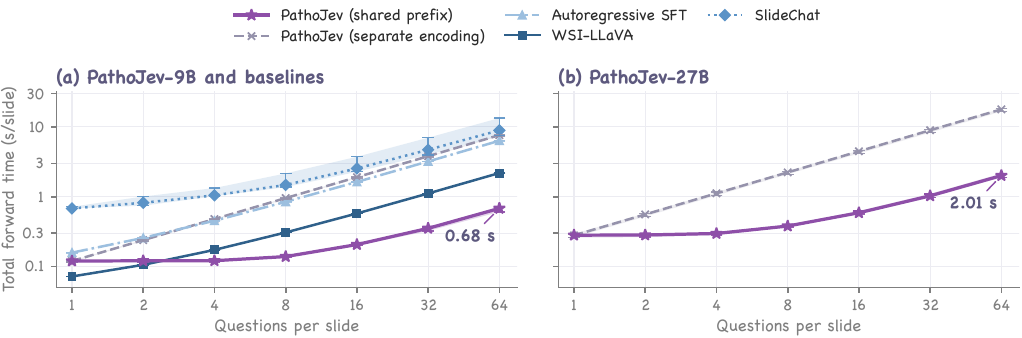}
    \caption{\textbf{Multi-question inference on WSI-Bench.} Median forward time over 40 slides; bands and error bars show 95\% bootstrap confidence intervals. Shared-prefix methods cache slide context; separate encoding repeats prefill. Questions are cycled beyond the available set. Offline feature extraction is excluded.}
    \label{fig:query_scaling}
\end{figure*}

\subsection{Analysis}
\label{sec:analysis}

\noindent\textbf{Question scaling and inference efficiency.}
Figure~\ref{fig:query_scaling} shows how total forward time scales with question count. At 64 questions, PathoJev-9B takes 0.68 seconds, versus 6.34 for autoregressive SFT, 2.18 for WSI-LLaVA, and 8.93 for SlideChat: speedups of $9.3\times$, $3.2\times$, and $13.1\times$ despite all three baselines reusing the slide cache. WSI-LLaVA, with a 576-token pooled prefix, is fastest at one and two questions; PathoJev-9B leads from four questions onward. The amortized cost falls from 17.3 to 10.6\,ms per question between $N=8$ and $N=64$ at 9B, and from 47.4 to 31.4\,ms at 27B. Thus, the one-time prefix cost in Eq.~\ref{eq:total_cost} is spread across more decisions while question suffixes are processed in batches. Appendix~\ref{app:efficiency_analysis} reports the measured workloads.

\begin{figure}[t]
    \centering
    \includegraphics[width=\columnwidth]{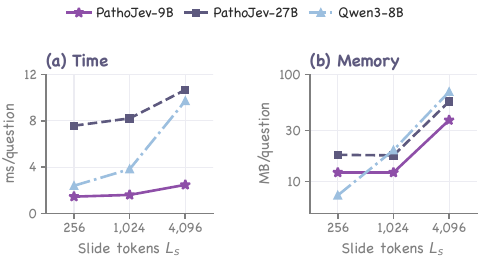}
    \caption{\textbf{Marginal cost of longer slide contexts.} Per-question time and memory increments are calculated from measurements at $N=1$ and $N=16$ after prefix prefill. Memory uses a log scale. All systems store one copy of the prefix keys and values.}
    \label{fig:context_scaling}
\end{figure}

\smallskip\noindent\textbf{Cost of longer slide context.}
Increasing slide context from 256 to 4,096 tokens raises marginal latency from 1.46 to 2.48\,ms per question at 9B and from 7.58 to 10.67\,ms at 27B (Figure~\ref{fig:context_scaling}): $1.7\times$ and $1.4\times$ increases for a $16\times$ longer context. The softmax-only Qwen3-8B increases from 2.40 to 9.72\,ms ($4.1\times$). All systems use the same shared-prefix attention implementation. Layer profiling follows Eq.~\ref{eq:row_cost}: over 16 question rows, 9B recurrent-layer time remains nearly constant at 58.1--58.3\,ms, while softmax-attention time rises from 19.0 to 33.8\,ms. Recurrent states carry the prefix summary into each question, leaving length-dependent work in softmax layers. At 4,096 tokens, marginal memory is 37.4\,MB for PathoJev-9B versus 68.8\,MB for Qwen3-8B, further reducing the cost of additional decisions.

\begin{table}[t]
    \centering
    \caption{\textbf{Calibration and accuracy at different coverage levels.} ECE uses 15 confidence bins at $\tau_t=1$ without post-hoc calibration. Coverage is the fraction of questions retained by confidence; $\Delta$ is the accuracy gain in percentage points.}
    \label{tab:calibration}
    \begingroup
    \footnotesize
    \setlength{\tabcolsep}{3pt}
    \renewcommand{\arraystretch}{1.18}
    \begin{tabularx}{\columnwidth}{>{\raggedright\arraybackslash}Xcc}
        \toprule
        \multicolumn{3}{l}{\textbf{(a) Calibration: ECE (percentage points, $\downarrow$)}} \\
        \midrule
        Benchmark & PathoJev-9B & PathoJev-27B \\
        \midrule
        WSI-VQA & \textbf{6.7} & \underline{10.5} \\
        WSI-Bench & \textbf{2.0} & \underline{2.1} \\
        SlideBench-TCGA & \textbf{2.2} & \underline{4.5} \\
        \bottomrule
    \end{tabularx}
    \par\smallskip
    \begin{tabularx}{\columnwidth}{>{\raggedright\arraybackslash}Xccc}
        \multicolumn{4}{l}{\textbf{(b) Accuracy by coverage: PathoJev-9B (\%)}} \\
        \midrule
        Benchmark & \makecell[c]{100\%\\coverage} & \makecell[c]{80\%\\coverage} & $\Delta$ \\
        \midrule
        WSI-VQA & 65.4 & 71.2 & +5.8 \\
        WSI-Bench & 90.4 & 97.0 & +6.6 \\
        SlideBench-TCGA & 75.9 & 83.2 & +7.3 \\
        \bottomrule
    \end{tabularx}
    \endgroup
\end{table}

\smallskip\noindent\textbf{Confidence and calibration.}
Retaining the 80\% most confident answers raises 9B accuracy from 90.4\% to 97.0\% on WSI-Bench and from 75.9\% to 83.2\% on SlideBench-TCGA (Table~\ref{tab:calibration}(b)); the latter reaches 93.2\% at 50\% coverage. Answer probabilities thus also support prioritizing less confident questions for further assessment. Without post-hoc calibration, WSI-Bench ECE is 2.0 and 2.1 percentage points at 9B and 27B, whereas WSI-VQA reaches 6.7 and 10.5 (Table~\ref{tab:calibration}(a)). Appendix~\ref{app:probability_metrics} provides ROI calibration and further coverage--accuracy results.

\smallskip\noindent\textbf{Image-conditioned pathology judgments.}
With the same grading question and answer set across 40 WSI-Bench slides, PathoJev-9B achieves 80.0\% accuracy, versus 40.0\% for SlideChat and 67.5\% for WSI-LLaVA (Figure~\ref{fig:qual_same_question}). CF-Path extends this comparison to binary statements. We form 117,377 positive--negative slide pairs with identical statement text within each TCGA project. The paired AUROC measures whether a supporting slide receives a higher true probability than an opposing slide according to the report-derived labels. PathoJev achieves 0.728 and 0.729 at 9B and 27B, versus 0.511 for the text-only control. Holding the question fixed across slides links both categorical answers and binary scores to the tissue context. Comparable paired scores at both scales show that this image dependence is already present at 9B. Construction and evaluation details appear in Appendices~\ref{app:training_data} and~\ref{app:visual_ablation}.

\section{Conclusion}
\label{sec:conclusion}

We presented PathoJev, a System One model for decisions over shared pathology context. It reads answer probabilities directly, while independent question branches reuse slide-prefix caches and fixed-size recurrent states. The same interface supports whole-slide and ROI assessment. Both scales surpass the evaluated pathology baselines on three whole-slide benchmarks. A same-data generative comparison and prefix-sharing ablations show complementary efficiency gains while retaining competitive accuracy across WSI and ROI tasks. These results provide a reusable probabilistic interface for pathology assessment across \mbox{whole-slide and ROI inputs}.

\iftoggle{cvprfinal}{}{\clearpage}
{
    \small
    \bibliographystyle{ieeenat_fullname}
    \bibliography{main}

\begin{thebibliography}{32}
\providecommand{\natexlab}[1]{#1}
\providecommand{\url}[1]{\texttt{#1}}
\expandafter\ifx\csname urlstyle\endcsname\relax
  \providecommand{\doi}[1]{doi: #1}\else
  \providecommand{\doi}{doi: \begingroup \urlstyle{rm}\Url}\fi

\bibitem[Almeida(2026)]{almeida2026introducing}
Diogo Almeida.
\newblock Introducing {System One} models \& {Jev}.
\newblock TypeSafe AI Blog, 2026.
\newblock Published September 15, 2026; accessed October 23, 2026.

\bibitem[{AutoTrust AI Lab}(2026{\natexlab{a}})]{autotrust2026jev27b}
{AutoTrust AI Lab}.
\newblock {JEV-27B-VL} model card.
\newblock Hugging Face, 2026{\natexlab{a}}.
\newblock Accessed: 2026-10-26.

\bibitem[{AutoTrust AI Lab}(2026{\natexlab{b}})]{autotrust2026jev9b}
{AutoTrust AI Lab}.
\newblock {JEV-9B} model card.
\newblock Hugging Face, 2026{\natexlab{b}}.
\newblock Accessed: 2026-10-26.

\bibitem[Chen et~al.(2024{\natexlab{a}})Chen, Gui, Ouyang, Gao, Chen, Chen,
  Wang, Cai, Ji, Wan, and Wang]{chen2024injecting}
Junying Chen, Chi Gui, Ruyi Ouyang, Anningzhe Gao, Shunian Chen, Guiming~Hardy
  Chen, Xidong Wang, Zhenyang Cai, Ke Ji, Xiang Wan, and Benyou Wang.
\newblock Towards injecting medical visual knowledge into multimodal {LLM}s at
  scale.
\newblock In \emph{Proceedings of the 2024 Conference on Empirical Methods in
  Natural Language Processing}, pages 7346--7370. Association for Computational
  Linguistics, 2024{\natexlab{a}}.

\bibitem[Chen et~al.(2024{\natexlab{b}})Chen, Li, Zhu, Zheng, Shui, and
  Yang]{chen2024wsicaption}
Pingyi Chen, Honglin Li, Chenglu Zhu, Sunyi Zheng, Zhongyi Shui, and Lin Yang.
\newblock {WsiCaption}: Multiple instance generation of pathology reports for
  gigapixel whole-slide images.
\newblock In \emph{Proceedings of Medical Image Computing and Computer Assisted
  Intervention -- MICCAI 2024}, pages 546--556. Springer Nature Switzerland,
  2024{\natexlab{b}}.

\bibitem[Chen et~al.(2025{\natexlab{a}})Chen, Zhu, Zheng, Li, and
  Yang]{chen2025wsivqa}
Pingyi Chen, Chenglu Zhu, Sunyi Zheng, Honglin Li, and Lin Yang.
\newblock {WSI-VQA}: Interpreting whole slide images by generative visual
  question answering.
\newblock In \emph{European Conference on Computer Vision}, pages 401--417.
  Springer, 2025{\natexlab{a}}.

\bibitem[Chen et~al.(2025{\natexlab{b}})Chen, Wang, Ji, Li, Ye, Li, Hu, Yu,
  Qiao, and He]{chen2025slidechat}
Ying Chen, Guoan Wang, Yuanfeng Ji, Yanjun Li, Jin Ye, Tianbin Li, Ming Hu,
  Rongshan Yu, Yu Qiao, and Junjun He.
\newblock {SlideChat}: A large vision-language assistant for whole-slide
  pathology image understanding.
\newblock In \emph{Proceedings of the IEEE/CVF Conference on Computer Vision
  and Pattern Recognition}, pages 5134--5143, 2025{\natexlab{b}}.

\bibitem[{College of American Pathologists}()]{cap2026protocols}
{College of American Pathologists}.
\newblock Cancer protocols frequently asked questions.
\newblock College of American Pathologists.
\newblock Accessed: 2026-10-23.

\bibitem[Ding et~al.(2025)Ding, Wagner, Song, Chen, Lu, Zhang, Vaidya, Jaume,
  Shaban, Kim, Williamson, Robertson, Chen, Almagro-P{\'e}rez, Doucet, Sahai,
  Chen, Chen, Komura, Kawabe, Ochi, Sato, Yokose, Miyagi, Ishikawa, Gerber,
  Peng, Le, and Mahmood]{ding2025multimodal}
Tong Ding, Sophia~J. Wagner, Andrew~H. Song, Richard~J. Chen, Ming~Y. Lu,
  Andrew Zhang, Anurag~J. Vaidya, Guillaume Jaume, Muhammad Shaban, Ahrong Kim,
  Drew F.~K. Williamson, Harry Robertson, Bowen Chen, Cristina
  Almagro-P{\'e}rez, Paul Doucet, Sharifa Sahai, Chengkuan Chen, Christina~S.
  Chen, Daisuke Komura, Akihiro Kawabe, Mieko Ochi, Shinya Sato, Tomoyuki
  Yokose, Yohei Miyagi, Shumpei Ishikawa, Georg Gerber, Tingying Peng, Long~Phi
  Le, and Faisal Mahmood.
\newblock A multimodal whole-slide foundation model for pathology.
\newblock \emph{Nature Medicine}, 31\penalty0 (11):\penalty0 3749--3761, 2025.

\bibitem[Guo et~al.(2017)Guo, Pleiss, Sun, and Weinberger]{guo2017calibration}
Chuan Guo, Geoff Pleiss, Yu Sun, and Kilian~Q. Weinberger.
\newblock On calibration of modern neural networks.
\newblock In \emph{Proceedings of the 34th International Conference on Machine
  Learning}, pages 1321--1330. PMLR, 2017.

\bibitem[Guo et~al.(2024)Guo, Ma, Xu, Wang, Wang, and Chen]{guo2024histgen}
Zhengrui Guo, Jiabo Ma, Yingxue Xu, Yihui Wang, Liansheng Wang, and Hao Chen.
\newblock {HistGen}: Histopathology report generation via local-global feature
  encoding and cross-modal context interaction.
\newblock In \emph{International Conference on Medical Image Computing and
  Computer-Assisted Intervention}, pages 189--199. Springer Nature Switzerland,
  2024.

\bibitem[He et~al.(2021)He, Cai, Wei, Zhang, Mou, Xing, and Xie]{he2021towards}
Xuehai He, Zhuo Cai, Wenlan Wei, Yichen Zhang, Luntian Mou, Eric Xing, and
  Pengtao Xie.
\newblock Towards visual question answering on pathology images.
\newblock In \emph{Proceedings of the 59th Annual Meeting of the Association
  for Computational Linguistics and the 11th International Joint Conference on
  Natural Language Processing (Volume 2: Short Papers)}, pages 708--718, 2021.

\bibitem[Hu et~al.(2022)Hu, Shen, Wallis, Allen-Zhu, Li, Wang, Wang, and
  Chen]{hu2022lora}
Edward~J. Hu, Yelong Shen, Phillip Wallis, Zeyuan Allen-Zhu, Yuanzhi Li, Shean
  Wang, Lu Wang, and Weizhu Chen.
\newblock {LoRA}: Low-rank adaptation of large language models.
\newblock In \emph{International Conference on Learning Representations}, 2022.

\bibitem[Hu et~al.(2024)Hu, Li, Lu, Shao, He, Qiao, and Luo]{hu2024omnimedvqa}
Yutao Hu, Tianbin Li, Quanfeng Lu, Wenqi Shao, Junjun He, Yu Qiao, and Ping
  Luo.
\newblock {OmniMedVQA}: A new large-scale comprehensive evaluation benchmark
  for medical {LVLM}.
\newblock In \emph{Proceedings of the IEEE/CVF Conference on Computer Vision
  and Pattern Recognition}, pages 22170--22183, 2024.

\bibitem[Huang et~al.(2023)Huang, Bianchi, Yuksekgonul, Montine, and
  Zou]{huang2023visual}
Zhi Huang, Federico Bianchi, Mert Yuksekgonul, Thomas~J. Montine, and James
  Zou.
\newblock A visual--language foundation model for pathology image analysis
  using medical {Twitter}.
\newblock \emph{Nature Medicine}, 29\penalty0 (9):\penalty0 2307--2316, 2023.

\bibitem[Liang et~al.(2025)Liang, Lyu, Chen, Ding, Zhang, He, Wu, Xing, Yang,
  Wang, and Shen]{liang2025wsi}
Yuci Liang, Xinheng Lyu, Wenting Chen, Meidan Ding, Jipeng Zhang, Xiangjian He,
  Song Wu, Xiaohan Xing, Sen Yang, Xiyue Wang, and Linlin Shen.
\newblock {WSI-LLaVA}: A multimodal large language model for whole slide image.
\newblock In \emph{Proceedings of the IEEE/CVF International Conference on
  Computer Vision}, pages 22718--22727, 2025.

\bibitem[Lu et~al.(2021)Lu, Williamson, Chen, Chen, Barbieri, and
  Mahmood]{lu2021data}
Ming~Y. Lu, Drew F.~K. Williamson, Tiffany~Y. Chen, Richard~J. Chen, Matteo
  Barbieri, and Faisal Mahmood.
\newblock Data-efficient and weakly supervised computational pathology on
  whole-slide images.
\newblock \emph{Nature Biomedical Engineering}, 5\penalty0 (6):\penalty0
  555--570, 2021.

\bibitem[Lu et~al.(2024{\natexlab{a}})Lu, Chen, Williamson, Chen, Liang, Ding,
  Jaume, Odintsov, Le, Gerber, Parwani, Zhang, and Mahmood]{lu2024visual}
Ming~Y. Lu, Bowen Chen, Drew F.~K. Williamson, Richard~J. Chen, Ivy Liang, Tong
  Ding, Guillaume Jaume, Igor Odintsov, Long~Phi Le, Georg Gerber, Anil~V.
  Parwani, Andrew Zhang, and Faisal Mahmood.
\newblock A visual-language foundation model for computational pathology.
\newblock \emph{Nature Medicine}, 30\penalty0 (3):\penalty0 863--874,
  2024{\natexlab{a}}.

\bibitem[Lu et~al.(2024{\natexlab{b}})Lu, Chen, Williamson, Chen, Zhao, Chow,
  Ikemura, Kim, Pouli, Patel, Soliman, Chen, Ding, Wang, Gerber, Liang, Le,
  Parwani, Weishaupt, and Mahmood]{lu2024multimodal}
Ming~Y. Lu, Bowen Chen, Drew F.~K. Williamson, Richard~J. Chen, Melissa Zhao,
  Aaron~K. Chow, Kenji Ikemura, Ahrong Kim, Dimitra Pouli, Ankush Patel, Amr
  Soliman, Chengkuan Chen, Tong Ding, Judy~J. Wang, Georg Gerber, Ivy Liang,
  Long~Phi Le, Anil~V. Parwani, Luca~L. Weishaupt, and Faisal Mahmood.
\newblock A multimodal generative {AI} copilot for human pathology.
\newblock \emph{Nature}, 634\penalty0 (8033):\penalty0 466--473,
  2024{\natexlab{b}}.

\bibitem[{Qwen Team}(2026{\natexlab{a}})]{qwen2026qwen35}
{Qwen Team}.
\newblock {Qwen3.5-9B} model card.
\newblock Hugging Face, 2026{\natexlab{a}}.
\newblock Accessed: 2026-10-26.

\bibitem[{Qwen Team}(2026{\natexlab{b}})]{qwen2026qwen38}
{Qwen Team}.
\newblock {Qwen3.8-27B} model card.
\newblock Hugging Face, 2026{\natexlab{b}}.
\newblock Accessed: 2026-10-26.

\bibitem[Schaad et~al.(2024)Schaad, Berezowska, Perren, and
  Hewer]{schaad2024impact}
Nicole Schaad, Sabina Berezowska, Aurel Perren, and Ekkehard Hewer.
\newblock Impact of template-based synoptic reporting on completeness of
  surgical pathology reports.
\newblock \emph{Virchows Archiv}, 484:\penalty0 31--36, 2024.

\bibitem[Seyfioglu et~al.(2024)Seyfioglu, Ikezogwo, Ghezloo, Krishna, and
  Shapiro]{seyfioglu2024quilt}
Mehmet~Saygin Seyfioglu, Wisdom~O. Ikezogwo, Fatemeh Ghezloo, Ranjay Krishna,
  and Linda Shapiro.
\newblock {Quilt-LLaVA}: Visual instruction tuning by extracting localized
  narratives from open-source histopathology videos.
\newblock In \emph{Proceedings of the IEEE/CVF Conference on Computer Vision
  and Pattern Recognition}, pages 13183--13192, 2024.

\bibitem[Shao et~al.(2021)Shao, Bian, Chen, Wang, Zhang, Ji, and
  Zhang]{shao2021transmil}
Zhuchen Shao, Hao Bian, Yang Chen, Yifeng Wang, Jian Zhang, Xiangyang Ji, and
  Yongbing Zhang.
\newblock {TransMIL}: Transformer based correlated multiple instance learning
  for whole slide image classification.
\newblock In \emph{Advances in Neural Information Processing Systems}, pages
  2136--2147, 2021.

\bibitem[Sun et~al.(2024)Sun, Wu, Zhu, Zheng, Chen, Zhang, Zhang, Wan, Lan,
  Zheng, Li, Lyu, Lin, and Yang]{sun2024pathmmu}
Yuxuan Sun, Hao Wu, Chenglu Zhu, Sunyi Zheng, Qizi Chen, Kai Zhang, Yunlong
  Zhang, Dan Wan, Xiaoxiao Lan, Mengyue Zheng, Jingxiong Li, Xinheng Lyu, Tao
  Lin, and Lin Yang.
\newblock {PathMMU}: A massive multimodal expert-level benchmark for
  understanding and reasoning in pathology.
\newblock In \emph{European Conference on Computer Vision}, pages 56--73.
  Springer, 2024.

\bibitem[Sun et~al.(2025)Sun, Zhang, Si, Zhu, Zhang, Shui, Li, Gong, Lyu, Lin,
  and Yang]{sun2025pathgen}
Yuxuan Sun, Yunlong Zhang, Yixuan Si, Chenglu Zhu, Kai Zhang, Zhongyi Shui,
  Jingxiong Li, Xuan Gong, Xinheng Lyu, Tao Lin, and Lin Yang.
\newblock {PathGen-1.6M}: 1.6 million pathology image-text pairs generation
  through multi-agent collaboration.
\newblock In \emph{International Conference on Learning Representations}, 2025.

\bibitem[Yang et~al.(2025)Yang, Kautz, and Hatamizadeh]{yang2025gated}
Songlin Yang, Jan Kautz, and Ali Hatamizadeh.
\newblock Gated delta networks: Improving {Mamba2} with delta rule.
\newblock In \emph{International Conference on Learning Representations}, 2025.

\bibitem[Yu and Yao(2026)]{yu2026visual}
Guanxu Yu and Yuhang Yao.
\newblock {Visual Jev}: Accurate and efficient decisions from shared visual
  context.
\newblock \emph{arXiv preprint arXiv:2609.25845}, 2026.

\bibitem[Zhang et~al.(2026)Zhang, Zhang, Guo, Cheng, Chen, Zhang, Zhang, Yi,
  and Bu]{zhang2026patho}
Wenchuan Zhang, Penghao Zhang, Jingru Guo, Tao Cheng, Jie Chen, Shuwan Zhang,
  Zhang Zhang, Yuhao Yi, and Hong Bu.
\newblock {Patho-R1}: A multimodal reinforcement learning-based pathology
  expert reasoner.
\newblock In \emph{Proceedings of the AAAI Conference on Artificial
  Intelligence}, pages 28418--28426, 2026.

\bibitem[Zhang et~al.(2024)Zhang, Wu, Zhao, Lin, Zhang, Wang, and
  Xie]{zhang2024development}
Xiaoman Zhang, Chaoyi Wu, Ziheng Zhao, Weixiong Lin, Ya Zhang, Yanfeng Wang,
  and Weidi Xie.
\newblock Development of a large-scale medical visual question-answering
  dataset.
\newblock \emph{Communications Medicine}, 4:\penalty0 277, 2024.

\bibitem[Zhu et~al.(2026)Zhu, Zeng, Si, Yang, Li, Qin, Zhou, Yao, Zhang, Qin,
  Zhang, Yang, Xu, Zhang, Guo, Wang, Duan, and Wang]{zhu2026videojev}
Yuhan Zhu, Xiangyu Zeng, Qingyi Si, Chenxu Yang, Yifei Li, Luozheng Qin, Junhao
  Zhou, Dingyu Yao, Jun Zhang, Chuanyu Qin, Zhiqiu Zhang, Yuandong Yang, Yuqi
  Xu, Tianyu Zhang, Hongcheng Guo, Jiaqi Wang, Nan Duan, and Limin Wang.
\newblock {VideoJev}: {System One} model for video understanding and
  interaction.
\newblock \emph{arXiv preprint arXiv:2610.18615}, 2026.

\bibitem[Zuo et~al.(2025)Zuo, Qu, Li, Chen, Zhu, Hua, Zhang, Ding, and
  Zhou]{zuo2025medxpertqa}
Yuxin Zuo, Shang Qu, Yifei Li, Zhang-Ren Chen, Xuekai Zhu, Ermo Hua, Kaiyan
  Zhang, Ning Ding, and Bowen Zhou.
\newblock {MedXpertQA}: Benchmarking expert-level medical reasoning and
  understanding.
\newblock In \emph{Proceedings of the 42nd International Conference on Machine
  Learning}, pages 80961--80990. PMLR, 2025.

\end{thebibliography}
}

\clearpage
\appendix
\maketitlesupplementary
\counterwithin{figure}{section}
\renewcommand{\thefigure}{\thesection\arabic{figure}}
\renewcommand{\theHfigure}{supp.\thesection.\arabic{figure}}
\setcounter{table}{0}
\renewcommand{\thetable}{A\arabic{table}}
\setcounter{algorithm}{0}
\renewcommand{\thealgorithm}{A\arabic{algorithm}}
\makeatletter
\providecommand{\theHALG@line}{}
\renewcommand{\theHALG@line}{\thealgorithm.\arabic{ALG@line}}
\makeatother

\section*{Supplementary Contents}
\begingroup
\small
\renewcommand{\arraystretch}{1.12}
\noindent\begin{tabular*}{\columnwidth}{@{}l@{\extracolsep{\fill}}r@{}}
\hyperref[app:experimental_details]{\ref*{app:experimental_details}\enspace \textbf{Training and Implementation}} & \pageref{app:experimental_details} \\
\hspace*{0.65em}\hyperref[app:optimization]{\ref*{app:optimization}\enspace Initialization and optimization} & \pageref{app:optimization} \\
\hspace*{0.65em}\hyperref[app:hardware]{\ref*{app:hardware}\enspace Hardware and checkpoints} & \pageref{app:hardware} \\
\hspace*{0.65em}\hyperref[app:training_data]{\ref*{app:training_data}\enspace Training data composition} & \pageref{app:training_data} \\
\hyperref[app:implementation]{\ref*{app:implementation}\enspace \textbf{Prompt Format and Algorithms}} & \pageref{app:implementation} \\
\hspace*{0.65em}\hyperref[app:prompt_format]{\ref*{app:prompt_format}\enspace Prompt serialization and readout} & \pageref{app:prompt_format} \\
\hspace*{0.65em}\hyperref[app:inference_algorithm]{\ref*{app:inference_algorithm}\enspace Shared-prefix inference} & \pageref{app:inference_algorithm} \\
\hspace*{0.65em}\hyperref[app:training_algorithm]{\ref*{app:training_algorithm}\enspace Two-stage adaptation} & \pageref{app:training_algorithm} \\
\hyperref[app:evaluation]{\ref*{app:evaluation}\enspace \textbf{Evaluation Protocols}} & \pageref{app:evaluation} \\
\hspace*{0.65em}\hyperref[app:benchmark_details]{\ref*{app:benchmark_details}\enspace Evaluation subsets} & \pageref{app:benchmark_details} \\
\hspace*{0.65em}\hyperref[app:baselines]{\ref*{app:baselines}\enspace Baseline models} & \pageref{app:baselines} \\
\hspace*{0.65em}\hyperref[app:latency_protocol]{\ref*{app:latency_protocol}\enspace Inference timing} & \pageref{app:latency_protocol} \\
\hspace*{0.65em}\hyperref[app:probability_metrics]{\ref*{app:probability_metrics}\enspace Probability evaluation} & \pageref{app:probability_metrics} \\
\hyperref[app:additional_ablations]{\ref*{app:additional_ablations}\enspace \textbf{Additional Ablation Results}} & \pageref{app:additional_ablations} \\
\hspace*{0.65em}\hyperref[app:readout_ablation]{\ref*{app:readout_ablation}\enspace Decision readout} & \pageref{app:readout_ablation} \\
\hspace*{0.65em}\hyperref[app:grid_resolution]{\ref*{app:grid_resolution}\enspace Slide grid resolution} & \pageref{app:grid_resolution} \\
\hspace*{0.65em}\hyperref[app:visual_ablation]{\ref*{app:visual_ablation}\enspace Visual input} & \pageref{app:visual_ablation} \\
\hyperref[app:efficiency_analysis]{\ref*{app:efficiency_analysis}\enspace \textbf{Inference Efficiency Analysis}} & \pageref{app:efficiency_analysis} \\
\hspace*{0.65em}\hyperref[app:query_scaling_details]{\ref*{app:query_scaling_details}\enspace Question-count scaling} & \pageref{app:query_scaling_details} \\
\hspace*{0.65em}\hyperref[app:marginal_cost]{\ref*{app:marginal_cost}\enspace Context and layer-wise costs} & \pageref{app:marginal_cost} \\
\hyperref[app:qualitative]{\ref*{app:qualitative}\enspace \textbf{Qualitative Examples}} & \pageref{app:qualitative} \\
\hspace*{0.65em}\hyperref[app:qual_selection]{\ref*{app:qual_selection}\enspace Case selection and display} & \pageref{app:qual_selection} \\
\hspace*{0.65em}\hyperref[app:qual_predictions]{\ref*{app:qual_predictions}\enspace Example predictions} & \pageref{app:qual_predictions} \\
\hyperref[app:future_work]{\ref*{app:future_work}\enspace \textbf{Limitations and Future Work}} & \pageref{app:future_work} \\
\end{tabular*}
\endgroup

\section{Training and Implementation}
\label{app:experimental_details}

\subsection{Initialization and Optimization}
\label{app:optimization}

PathoJev-9B and PathoJev-27B are initialized from JEV-9B and JEV-27B-VL~\cite{autotrust2026jev9b,autotrust2026jev27b}, respectively, with Qwen3.5-9B and Qwen3.8-27B backbones~\cite{qwen2026qwen35,qwen2026qwen38}. We retain their pretrained LoRA adapters and typed decision heads. Table~\ref{tab:training_hyperparameters} summarizes the two-stage schedule shared by both scales. Stage A aligns the slide projection while the decision model is fixed; Stage B adapts the projection, LoRA adapters, and head jointly using WSI and ROI questions.

Both stages use AdamW with weight decay 0.01, gradient clipping at 1.0, PyTorch's default moment and epsilon parameters, and seed 0. LoRA uses rank $r=16$, scaling $\alpha=32$, and dropout 0.05. It targets the language model's query, key, value, and output projections, the feed-forward gate, up, and down projections, and the Gated DeltaNet projections \texttt{in\_proj\_qkv}, \texttt{in\_proj\_z}, and \texttt{out\_proj}. No adapters are added to the vision encoder.

\begin{table}[htbp]
    \centering
    \caption{Two-stage training schedule, shared by the 9B and 27B models. Question counts are per optimization step.}
    \label{tab:training_hyperparameters}
    \footnotesize
    \setlength{\tabcolsep}{3pt}
    \renewcommand{\arraystretch}{1.15}
    \begin{tabularx}{\columnwidth}{>{\raggedright\arraybackslash}Xcc}
        \toprule
        Setting & Stage A & Stage B \\
        \midrule
        Trainable parameters & $\phi$ & $\phi,\Delta,\psi$ \\
        Training slides used & 30\% & 100\% \\
        Epochs & 1 & 2 \\
        Optimization steps & 244 & 1,624 \\
        Slides per step & 8 & 8 \\
        Questions per slide & $\leq 32$ & $\leq 64$ \\
        ROI questions per step & 0 & 32 \\
        \midrule
        Projection learning rate & $10^{-3}$ & $2\times10^{-4}$ \\
        LoRA learning rate & Frozen & $10^{-4}$ \\
        Head learning rate & Frozen & $10^{-4}$ \\
        Linear warm-up steps & 100 & 48 \\
        Post-warm-up schedule & Constant & Cosine to zero \\
        \midrule
        Optimizer & \multicolumn{2}{c}{AdamW} \\
        Weight decay & \multicolumn{2}{c}{0.01} \\
        Gradient clipping & \multicolumn{2}{c}{1.0} \\
        LoRA rank / scaling & \multicolumn{2}{c}{16 / 32} \\
        LoRA dropout & \multicolumn{2}{c}{0.05} \\
        Seed & \multicolumn{2}{c}{0} \\
        \bottomrule
    \end{tabularx}
\end{table}

\paragraph{Decision temperature.}
We fix $\tau_t=1$ for all output types throughout training and evaluation and apply no post-hoc calibration. All ECE and coverage--accuracy results use these uncalibrated probabilities.

\subsection{Hardware and Checkpoint Selection}
\label{app:hardware}

Training uses four RTX PRO 6000 GPUs with 96\,GB memory each, with data parallelism and gradient all-reduce.

For 9B, validation runs every 250 Stage B steps score checkpoints by the mean of PathMMU and WSI-VQA validation accuracy and CF-Path validation paired AUROC. This rule selects the checkpoint at Stage B step 1,250; for 27B, we use the final checkpoint. WSI question rows are processed in chunks of 16 at 27B, each reading the same slide prefix with isolated question contexts. ROI microbatches contain 32 questions at 9B and 16 at 27B. Chunking preserves the per-step sample counts and training objective.

\subsection{Training Data Composition}
\label{app:training_data}

Table~\ref{tab:training_data} lists the question counts in the training mixture. Whole-slide questions come from the training portions of SlideInstruct, WSI-Bench, and WSI-VQA, together with CF-Path statements constructed from TCGA reports. ROI supervision comes from PathVQA and the pathology subset of PMC-VQA. In Stage A, we sample 30\% of the training slides. Stage B uses the full training mixture, including the slides used in Stage A.

\begin{table}[htbp]
    \centering
    \caption{Training mixture used for pathology adaptation.}
    \label{tab:training_data}
    \small
    \begin{tabular}{llr}
        \toprule
        Source & Input & Questions \\
        \midrule
        SlideInstruct & WSI & 48,699 \\
        WSI-Bench & WSI & 34,799 \\
        WSI-VQA & WSI & 2,311 \\
        CF-Path & WSI & 126,467 \\
        PathVQA & ROI & 9,751 \\
        PMC-VQA (pathology) & ROI & 8,204 \\
        \bottomrule
    \end{tabular}
\end{table}

\paragraph{CF-Path construction.}
CF-Path uses diagnostic slides and their PathText report summaries~\cite{chen2024wsicaption} from 11 TCGA projects: BLCA, BRCA, COAD, HNSC, KIRC, LUAD, LUSC, PRAD, READ, STAD, and UCEC. Qwen3.5-9B in text-only, non-thinking mode decomposes each report into atomic statements, assigns a category to each statement, and generates a minimally modified false version. Category filtering retains diagnosis, grade, morphology, invasion, tissue type, and lymph-node metastasis pertaining to the current slide. Statements about laterality, gross findings or dimensions, and lymph nodes outside the current slide are excluded.

Statements from the case's own report receive positive labels, and their perturbed versions receive negative labels. Two additional sources create statements shared across cases within the same TCGA project: 12 statements from other cases and the 60 most frequent eligible statements, each occurring in at least three cases. Qwen3.5-9B assesses these statements against the target case's report using natural language inference. Entailment probability of at least 0.8 yields a positive label; contradiction probability of at least 0.8 yields a negative label. Remaining candidates are discarded. The resulting binary labels are silver-standard annotations produced automatically from reports, without statement-level manual verification.

\paragraph{CF-Path splits and training exclusions.}
We group cases by TCGA tissue source site (TSS) code and stratify by project, targeting a 70/15/15 training/validation/test split. Assignments for the initial five projects remain fixed when additional projects are incorporated; only previously unassigned TSS groups receive new assignments. Cases and TSS groups do not cross splits. Table~\ref{tab:cfpath_splits} summarizes the resulting sets. For PathoJev adaptation, we remove all training rows whose case appears in the WSI-VQA, WSI-Bench, or SlideBench evaluation sets, or in CF-Path validation or test sets.

\begin{table}[htbp]
    \centering
    \caption{CF-Path split sizes. TSS denotes tissue source site.}
    \label{tab:cfpath_splits}
    \footnotesize
    \begin{tabular}{lrrr}
        \toprule
        & Training & Validation & Test \\
        \midrule
        Questions & 126,467 & 36,372 & 38,088 \\
        Slides & 2,926 & 627 & 628 \\
        Cases & 2,721 & 595 & 623 \\
        TSS groups & 111 & 66 & 76 \\
        \bottomrule
    \end{tabular}
\end{table}

\FloatBarrier
\section{Prompt Format and Algorithms}
\label{app:implementation}

\subsection{Prompt Serialization and Decision Readout}
\label{app:prompt_format}

PathoJev uses JEV's \texttt{bare-v1} format. The markers \texttt{[state]}, \texttt{[kind]}, \texttt{[question]}, \texttt{[options]}, and \texttt{[decision]:} are lowercase ordinary text processed by the existing tokenizer. The boxes below show the field order; braces indicate question-specific content, and \texttt{<V>} denotes inserted slide embeddings.

\begin{promptbox}{Whole-slide multiple-choice prompt}
[state] Whole-slide image (H&E): <V>
[kind] choice
[question] {question}
[options]
A) {option A}
B) {option B}
...
[decision]:
\end{promptbox}

The WSI prefix contains a trailing space after the colon, the projected slide embeddings, and a newline. It is encoded once. Each question suffix begins with \texttt{[kind]} and branches independently from the prefix cache. Multiple-choice alternatives are labeled A through P, with only the available alternatives included. Binary questions use the following suffix and preserve the order \texttt{false}, then \texttt{true}.

\begin{promptbox}{Binary question suffix}
[kind] noul
[question] {statement}
[options]
false
true
[decision]:
\end{promptbox}

For ROI inputs, \texttt{[kind]} precedes \texttt{[state]}. Each image is processed with one question through the native visual pathway. The state line contains the image placeholders used by the backbone's processor.

\begin{promptbox}{ROI multiple-choice prompt}
[kind] choice
[state] An H&E-stained histopathology image: <|vision_start|><|image_pad|><|vision_end|>
[question] {question}
[options]
A) {option A}
B) {option B}
...
[decision]:
\end{promptbox}

The typed head reads the final normalized hidden state at the last token of the question. With the Qwen3.5-9B tokenizer, this is the token containing \texttt{]:} at the end of \texttt{[decision]:}. PathoJev applies the head directly to this bare prompt; the autoregressive control uses the Qwen chat template, and other generative baselines use their respective model templates.

\subsection{Shared-Prefix Decision Inference}
\label{app:inference_algorithm}

Algorithm~\ref{alg:shared_inference} summarizes WSI inference. Each recurrent layer initializes a separate branch state for every question, while prefix attention keys and values are shared. Suffix states and keys and values remain local to each row, so batching changes execution order without changing question context. The decision head selects the output indices associated with the question type and normalizes over its valid candidates.

\begin{algorithm}[tbp]
    \caption{Shared-prefix whole-slide decisions}
    \label{alg:shared_inference}
    \footnotesize
    \begin{algorithmic}[1]
        \Require Patch features $\{f_j,r_j\}_{j=1}^{M}$; questions $\{q_i,t_i,\mathcal{O}_i\}_{i=1}^{N}$; parameters $\theta,\phi,W_d,b_d$
        \Ensure Answer distributions $\{p_i\}$ and predictions $\{\hat y_i\}$
        \State Compute $V_X$ by spatial mean pooling and projection (Eq.~\ref{eq:slide_encoding})
        \State $P_X \gets E(c_{\mathrm{WSI}})\Vert V_X\Vert E(\texttt{\textbackslash n})$
        \State $\mathcal{C}_X \gets \operatorname{Prefill}_{\theta}(P_X)$
        \For{each batch $\mathcal{I}$ of question indices}
            \For{each $i\in\mathcal{I}$}
                \State Serialize $R_i$ using Eq.~\ref{eq:typed_prompt}
                \State Initialize row-local recurrent and convolutional states from $\mathcal{C}_X$
                \State Share prefix keys and values from $\mathcal{C}_X$
            \EndFor
            \State Evaluate rows $\{R_i\}_{i\in\mathcal{I}}$ independently in a batch
            \For{each $i\in\mathcal{I}$}
                \State $h_i \gets$ final normalized state at \texttt{[decision]:}
                \State $z_i \gets W_dh_i+b_d$
                \State $u_i \gets (z_{i,a_{t_i}(1)},\ldots,z_{i,a_{t_i}(K_i)})$
                \State $p_i \gets \operatorname{softmax}(u_i)$ \Comment{$\tau_{t_i}=1$}
                \State $\hat y_i \gets \arg\max_k p_{ik}$
            \EndFor
        \EndFor
        \State \Return $\{p_i,\hat y_i\}_{i=1}^{N}$
    \end{algorithmic}
\end{algorithm}

\subsection{Two-Stage Adaptation}
\label{app:training_algorithm}

Algorithm~\ref{alg:adaptation} summarizes the optimization in Sec.~\ref{sec:training_recipe}. WSI questions use the shared-prefix layout; ROI questions use the native visual pathway. During training, prefix computation remains in the gradient graph so that decision supervision can update the WSI projection. The stage-specific sampling and learning-rate schedules are given in Table~\ref{tab:training_hyperparameters}.

\begin{algorithm}[tbp]
    \caption{Two-stage pathology adaptation}
    \label{alg:adaptation}
    \footnotesize
    \begin{algorithmic}[1]
        \Require Pretrained backbone $\theta_0$, adapters $\Delta_0$, head $\psi_0$; stage-specific question distributions $\mathcal{D}_A,\mathcal{D}_B$
        \State Initialize projection $\phi$; set $\Delta\gets\Delta_0$, $\psi\gets\psi_0$
        \State Freeze $\theta_0$ and both visual encoders
        \For{$s\in\{A,B\}$}
            \If{$s=A$}
                \State $\Omega\gets\{\phi\}$; freeze $\Delta,\psi$; $S\gets244$
            \Else
                \State $\Omega\gets\{\phi,\Delta,\psi\}$; unfreeze $\Delta,\psi$; $S\gets1{,}624$
            \EndIf
            \State Initialize AdamW over $\Omega$ with the Stage $s$ schedule
            \For{$j=1,\ldots,S$}
                \State Sample WSI questions from $\mathcal{D}_s$
                \State Include ROI questions in Stage B
                \State Predict $p_{ik}$ using the appropriate visual pathway
                \State Compute $\mathcal{L}_{\mathcal{D}_s}$ from $-\log p_{i,y_i}$ (Eq.~\ref{eq:training_risk})
                \State Backpropagate to $\Omega$ and clip gradients at 1.0
                \State Update $\Omega$ using the learning rates at step $j$
            \EndFor
        \EndFor
        \State Select the checkpoint as in Appendix~\ref{app:hardware}
    \end{algorithmic}
\end{algorithm}

\FloatBarrier
\section{Evaluation Protocols}
\label{app:evaluation}

\subsection{Evaluation Subsets}
\label{app:benchmark_details}

Table~\ref{tab:evaluation_subsets} gives the subsets used for the eight-benchmark evaluation. WSI-VQA contains 85 slides, WSI-Bench contains 360, and SlideBench-TCGA contains 1,269. Accuracy is computed over questions within each benchmark. WSI and ROI means assign equal weight to each benchmark in the corresponding group. Figure~\ref{fig:overview} averages each model's available benchmark results within that group.

\begin{table}[htbp]
    \centering
    \caption{Evaluation subsets and question counts.}
    \label{tab:evaluation_subsets}
    \small
    \begin{tabular}{lr}
        \toprule
        Benchmark / subset & Questions \\
        \midrule
        \multicolumn{2}{l}{\textit{Whole-slide assessment}} \\
        WSI-VQA & 390 \\
        WSI-Bench (closed-ended) & 1,141 \\
        SlideBench-TCGA & 3,127 \\
        \midrule
        \multicolumn{2}{l}{\textit{Local tissue assessment}} \\
        PathMMU-tiny & 1,139 \\
        PathVQA (yes/no) & 3,362 \\
        Quilt-VQA (closed-ended) & 343 \\
        OmniMedVQA (pathology) & 890 \\
        MedXpertQA (pathology) & 90 \\
        \bottomrule
    \end{tabular}
\end{table}

\subsection{Baseline Models}
\label{app:baselines}
We compare with Qwen3.5-9B and Qwen3.8-27B in generation mode, their zero-shot JEV counterparts, and Thinking-mode configurations at both scales. Specialized baselines include Patho-R1-7B~\cite{zhang2026patho}, HuatuoGPT-Vision-7B~\cite{chen2024injecting}, Quilt-LLaVA-v1.5-7B~\cite{seyfioglu2024quilt}, and PathGen-LLaVA-13B~\cite{sun2025pathgen}, together with our reproductions of SlideChat~\cite{chen2025slidechat} and WSI-LLaVA~\cite{liang2025wsi}. Zero-shot WSI baselines receive slide thumbnails, while PathoJev uses the patch-feature pathway in Sec.~\ref{sec:decision_framework}. PathoJev uses the SlideInstruct and WSI-Bench training data also used by SlideChat and WSI-LLaVA, respectively.

The thumbnail-based pathology VLMs use their own conversation templates and greedy decoding with a 32-token output limit. HuatuoGPT-Vision, Quilt-LLaVA, and PathGen-LLaVA are evaluated on all three WSI test sets; unparseable answers are counted as incorrect.

\paragraph{Thinking-mode decoding.}
Both Qwen scales use temperature 1.0, top-$p$ 0.95, top-$k$ 20, and presence penalty 1.5. The thinking budget is 8,192 tokens. When that budget is reached, the thinking segment is closed and at most 512 additional tokens are allowed for the final answer. Seeds are assigned by global question index, and unparseable answers count as incorrect. The latency runs use batch size one.

\subsection{Inference Timing Protocol}
\label{app:latency_protocol}

We measure GPU forward time after untimed warm-up, with GPU synchronization around the timed computation. Measurements use an exclusively allocated RTX PRO 6000 GPU with 96\,GB memory and batch size one. File loading and input preprocessing are outside the timed region.

\paragraph{Whole-slide inputs.}
PathoJev, SlideChat, and WSI-LLaVA start from pre-extracted features already resident on the GPU. CONCH or GigaPath patch-feature extraction is excluded. For PathoJev, timing includes the WSI projection MLP, shared-prefix prefill, and question-row forward passes. Mean pooling onto the $32\times32$ grid is performed offline. For SlideChat, timing includes its LongNet slide encoder, projection, prefix prefill, and greedy decoding for each question from a copy of the prefix cache. For WSI-LLaVA, timing includes prefix prefill and decoding. Its Prov-GigaPath LongNet slide encoder and pooling to 576 visual tokens run during feature preparation and are excluded from the measured latency.

\paragraph{ROI inputs.}
The timed region covers the model's generation or forward call, including visual encoding, prefix prefill, and answer decoding or decision readout. Image decoding and processor operations run before timing begins.

\paragraph{Thumbnail-based WSI inputs.}
Qwen, JEV, and the zero-shot pathology VLMs receive a slide thumbnail separately for each question. Their timed region includes visual encoding, prefix prefill, and answer decoding or decision readout, as for ROI inputs. The three additional pathology VLM runs use five untimed warm-up questions and all 1,141 WSI-Bench questions.

\paragraph{Thinking-mode timing subsets.}
Thinking-mode latency is the mean over the first 100 questions in test-file order on PathMMU-tiny or WSI-Bench, at each model scale. This timing subset is separate from the accuracy evaluation. The 9B WSI mean is 82.3 seconds per question (median 61.5); the 27B ROI and WSI means are 78.6 and 69.1 (medians 38.9 and 40.9). Table~\ref{tab:main_results} reports the means.

\subsection{Probability Evaluation}
\label{app:probability_metrics}

For each question, confidence is the largest predicted answer probability. Expected calibration error~\cite{guo2017calibration} aggregates the absolute difference between accuracy and mean confidence across 15 confidence bins, weighted by the number of questions in each bin. Coverage--accuracy curves rank questions by confidence and report accuracy on the retained fraction. Risk is the error rate among retained questions, and AURC summarizes risk over coverage, with lower values indicating lower error across coverage levels. These metrics use the probabilities obtained at $\tau_t=1$, as specified in Appendix~\ref{app:optimization}.

Table~\ref{tab:calibration}(a) reports whole-slide ECE for both model scales, and Table~\ref{tab:calibration}(b) reports PathoJev-9B accuracy when retaining the most confident 80\% of questions. At 50\% coverage, SlideBench-TCGA accuracy reaches 93.2\%. On ROI inputs, the 9B model has ECEs of 14.0 and 18.1 percentage points on PathMMU-tiny and Quilt-VQA, respectively.

\FloatBarrier
\section{Additional Ablation Results}
\label{app:additional_ablations}

\subsection{Decision Readout Ablation}
\label{app:readout_ablation}

The autoregressive control uses the Qwen3.5-9B backbone and follows PathoJev's two-stage schedule, pathology training data, and step counts. It starts from Qwen3.5-9B weights, whereas PathoJev starts from JEV-9B (Appendix~\ref{app:optimization}). The control learns textual answers with the Qwen chat template, whereas PathoJev uses the bare prompt and typed decision head described in Appendix~\ref{app:prompt_format}. Its checkpoint is selected with the validation rule in Appendix~\ref{app:hardware}, at Stage B step 750. Greedy decoding is limited to 16 output tokens. The WSI generative control reuses the slide-prefix cache. Table~\ref{tab:readout_ci} reports paired-bootstrap 95\% confidence intervals for accuracy differences. Per-benchmark differences are computed before rounding the displayed accuracies.

\begin{table}[htbp]
    \centering
    \caption{Accuracy differences between PathoJev-9B and autoregressive SFT, in percentage points, with paired-bootstrap 95\% confidence intervals.}
    \label{tab:readout_ci}
    \footnotesize
    \begin{tabular}{lrrr}
        \toprule
        Benchmark & $\Delta$ & Lower & Upper \\
        \midrule
        WSI-VQA & $+1.5$ & $-1.8$ & $+4.9$ \\
        WSI-Bench & $+1.4$ & $0.0$ & $+2.8$ \\
        SlideBench-TCGA & $+1.7$ & $+0.5$ & $+2.9$ \\
        \midrule
        PathMMU-tiny & $-2.4$ & $-4.5$ & $-0.3$ \\
        PathVQA & $+0.8$ & $+0.1$ & $+1.5$ \\
        Quilt-VQA & $+4.1$ & $0.0$ & $+8.2$ \\
        OmniMedVQA & $+6.5$ & $+3.9$ & $+9.1$ \\
        MedXpertQA & $-4.4$ & $-11.1$ & $+2.2$ \\
        \bottomrule
    \end{tabular}
\end{table}

The SlideBench-TCGA interval is positive, while the WSI-VQA interval includes zero. The WSI-Bench interval touches zero at the reported precision. ROI intervals show gains on PathVQA and OmniMedVQA and a decrease on PathMMU-tiny; the Quilt-VQA interval touches zero and the MedXpertQA interval includes zero.

\subsection{Slide Grid Resolution}
\label{app:grid_resolution}

Table~\ref{tab:grid_resolution} compares $32\times32$ and $64\times64$ grids under the initial training-data configuration. Both runs start from JEV-9B and use the same training data, learning-rate schedule, seed 0, and step counts: 125 in Stage A and 831 in Stage B. Each step uses eight slides and 32 ROI questions. The row and column position-embedding tables match the grid size. Stage A uses up to 32 questions per slide for the $32\times32$ run and up to 64 for the $64\times64$ run. The runs use two GPUs and one GPU, respectively, with the same per-step sample counts.

Every 250 Stage B steps, we select checkpoints using the mean of PathMMU and WSI-VQA validation accuracy and CF-Path validation paired AUROC. The selected checkpoints are steps 500 and 250 for the $32\times32$ and $64\times64$ runs, respectively. Over 300 sampled slides, median visual sequence length increases from 499 to 1,755 tokens. The finer-grid run has lower accuracy on all three WSI benchmarks, with a $3.5\times$ larger median token count. These results describe the two training configurations above; Table~\ref{tab:main_results} uses the final training configuration.

\begin{table}[htbp]
    \centering
    \caption{Grid-resolution comparison under the initial training-data configuration. Accuracy (\%); visual token counts are medians over 300 sampled slides.}
    \label{tab:grid_resolution}
    \footnotesize
    \begin{tabular}{lrr}
        \toprule
        Measure & $32\times32$ & $64\times64$ \\
        \midrule
        Median visual tokens & 499 & 1,755 \\
        \midrule
        WSI-VQA & 67.7 & 65.9 \\
        WSI-Bench & 88.0 & 84.7 \\
        SlideBench-TCGA & 73.6 & 68.8 \\
        \midrule
        PathMMU-tiny & 75.9 & 74.4 \\
        PathVQA & 93.4 & 93.0 \\
        Quilt-VQA & 68.2 & 66.2 \\
        OmniMedVQA & 75.3 & 77.8 \\
        \bottomrule
    \end{tabular}
\end{table}

\subsection{Visual Input Ablation}
\label{app:visual_ablation}

Table~\ref{tab:visual_ablation} compares image-conditioned predictions with predictions made without images. Questions and answer alternatives are retained. Without the image, the 9B model loses accuracy on seven of eight benchmarks, with drops ranging from 6.7 to 35.5 percentage points. On MedXpertQA (90 clinical-vignette questions), accuracy is 25.6\% with images and 28.9\% without them. The 27B model retains 86.2\% accuracy on PathVQA without an image, compared with 95.1\% with an image, indicating substantial predictive information in the textual input on this benchmark.

\begin{table}[htbp]
    \centering
    \caption{Visual input ablation. Accuracy (\%) with and without the image, and the decrease after image removal in percentage points.}
    \label{tab:visual_ablation}
    \footnotesize
    \begin{tabular}{lrrr}
        \toprule
        Benchmark & With image & No image & Decrease \\
        \midrule
        \multicolumn{4}{l}{\textcolor{black!65}{\textbf{PathoJev-9B}}} \\
        WSI-VQA & 65.4 & 58.7 & 6.7 \\
        WSI-Bench & 90.4 & 80.3 & 10.1 \\
        SlideBench-TCGA & 75.9 & 63.9 & 12.0 \\
        PathMMU-tiny & 71.9 & 54.5 & 17.4 \\
        PathVQA & 94.6 & 62.1 & 32.5 \\
        Quilt-VQA & 70.8 & 49.6 & 21.2 \\
        OmniMedVQA & 75.3 & 39.8 & 35.5 \\
        \midrule
        \multicolumn{4}{l}{\textcolor{black!65}{\textbf{PathoJev-27B}}} \\
        PathVQA & 95.1 & 86.2 & 8.9 \\
        \bottomrule
    \end{tabular}
\end{table}

\paragraph{Paired-statement diagnostic.}
For CF-Path evaluation, we group statements with identical normalized text within each TCGA project and form pairs with opposite binary labels. The test set contains 38,088 questions and 117,377 such positive--negative pairs. PathoJev-9B and PathoJev-27B achieve within-statement paired AUROCs of 0.728 and 0.729, respectively, compared with 0.511 for the text-only control. This diagnostic holds statement content fixed across slides and measures agreement with the report-derived labels described in Appendix~\ref{app:training_data}.

\FloatBarrier
\section{Inference Efficiency Analysis}
\label{app:efficiency_analysis}

\subsection{Question-Count Scaling}
\label{app:query_scaling_details}
Figure~\ref{fig:query_scaling} reports median total forward time over the same 40 WSI-Bench slides at $N\in\{1,2,4,8,16,32,64\}$. Table~\ref{tab:query_scaling} gives representative values. Pointwise 95\% confidence intervals use 10,000 bootstrap resamples of slides with replacement (seed 0), preserving all question counts and model settings from each resampled slide. The public WSI benchmarks contain an average of 2.5--4.6 questions per slide, with at most eight. We cycle through each slide's own questions to construct larger workloads. The generative baselines and shared-prefix PathoJev reuse the prefix; the separate-encoding variant recomputes it for each question. Timing follows Appendix~\ref{app:latency_protocol}. The matched comparison with separate prefix encoding is given in Table~\ref{tab:prefix_ablation}.

\begin{table}[htbp]
    \centering
    \caption{Median total forward time (seconds per slide, $\downarrow$) over 40 WSI-Bench slides. All models reuse the slide-prefix cache.}
    \label{tab:query_scaling}
    \footnotesize
    \renewcommand{\arraystretch}{1.18}
    \begin{tabularx}{\columnwidth}{>{\raggedright\arraybackslash}Xccc}
        \toprule
        Model & $N=1$ & $N=8$ & $N=64$ \\
        \midrule
        Autoregressive SFT & 0.16 & 0.84 & 6.34 \\
        WSI-LLaVA & \textbf{0.07} & \underline{0.31} & \underline{2.18} \\
        SlideChat & 0.68 & 1.47 & 8.93 \\
        \rowcolor{blue!10}
        PathoJev-9B & \underline{0.12} & \textbf{0.14} & \textbf{0.68} \\
        \bottomrule
    \end{tabularx}
\end{table}

\begin{figure}[htbp]
    \centering
    \includegraphics[width=\columnwidth]{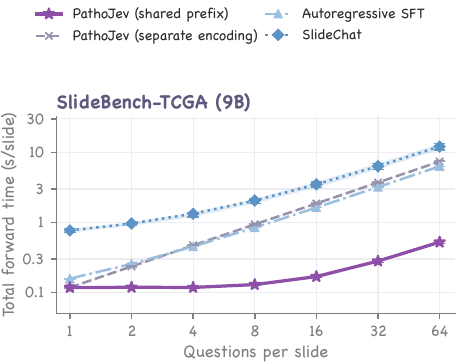}
    \caption{\textbf{Question scaling on SlideBench-TCGA.} Median total forward time over 40 slides; shading and capped error bars show pointwise 95\% bootstrap confidence intervals. Timing and repeated-question construction follow the WSI-Bench analysis.}
    \label{fig:slidebench_scaling}
\end{figure}

Figure~\ref{fig:slidebench_scaling} shows the corresponding 9B analysis on 40 SlideBench-TCGA slides. At 64 questions, shared-prefix PathoJev takes 0.52 seconds, compared with 7.39 for separate encoding, 6.32 for autoregressive SFT, and 12.12 for SlideChat.

\subsection{Context-Length and Layer-Wise Costs}
\label{app:marginal_cost}
\label{app:layer_timing}
After prefix prefill, we measure row execution time $T(N)$ and peak GPU memory $M(N)$. Marginal costs are $[T(16)-T(1)]/15$ and $[M(16)-M(1)]/15$, with memory in decimal MB. Table~\ref{tab:context_cost} supplements Figure~\ref{fig:context_scaling} with exact values and layer timings. All systems use the same shared-prefix attention implementation, storing one copy of the prefix keys and values. The comparison measures complete models with different layer counts, dimensions, and recurrent-state sizes.

Recurrent-layer time remains stable as $L_s$ grows, while softmax-attention time increases. The former reuses fixed-size states; the latter reads a prefix cache whose length grows with $L_s$ (Sec.~\ref{sec:shared_context}).

At $L_s=256$, Qwen3-8B has lower marginal memory than PathoJev-9B (7.4 versus 12.1\,MB per question). PathoJev initializes a separate fixed-size recurrent state for each row. As context length increases to 4,096, its marginal memory grows to 37.4\,MB, compared with 68.8\,MB for Qwen3-8B. The memory advantage therefore emerges at longer contexts rather than holding across all measured lengths.

\begin{table}[!htbp]
    \centering
    \caption{Costs at three slide-context lengths $L_s$ with one shared prefix KV copy in every system. Marginal costs are per question; layer timings aggregate 16 rows.}
    \label{tab:context_cost}
    \label{tab:layer_cost}
    \footnotesize
    \renewcommand{\arraystretch}{1.1}
    \begin{tabularx}{\columnwidth}{>{\raggedright\arraybackslash}Xrrr}
        \toprule
        Measure & 256 & 1,024 & 4,096 \\
        \midrule
        \rowcolor{blue!10}
        \multicolumn{4}{l}{\textbf{PathoJev-9B}} \\
        Marginal time (ms/question) & \textbf{1.46} & \textbf{1.61} & \textbf{2.48} \\
        Marginal memory (MB/question) & \underline{12.1} & \textbf{12.1} & \textbf{37.4} \\
        Recurrent layers (ms) & 58.1 & 58.2 & 58.3 \\
        Softmax attention (ms) & 19.0 & 20.9 & 33.8 \\
        \midrule
        \rowcolor{blue!10}
        \multicolumn{4}{l}{\textbf{PathoJev-27B}} \\
        Marginal time (ms/question) & 7.58 & 8.21 & 10.67 \\
        Marginal memory (MB/question) & 17.7 & \underline{17.5} & \underline{55.7} \\
        Recurrent layers (ms) & 174.5 & 174.3 & 173.9 \\
        Softmax attention (ms) & 55.4 & 61.3 & 99.9 \\
        \midrule
        \multicolumn{4}{l}{\textbf{Qwen3-8B}} \\
        Marginal time (ms/question) & \underline{2.40} & \underline{3.83} & \underline{9.72} \\
        Marginal memory (MB/question) & \textbf{7.4} & 19.6 & 68.8 \\
        Softmax attention (ms) & 59.5 & 80.8 & 173.6 \\
        \bottomrule
    \end{tabularx}
\end{table}

\FloatBarrier
\section{Qualitative Examples}
\label{app:qualitative}

\begin{figure*}[tp]
    \centering
    \includegraphics[width=\textwidth]{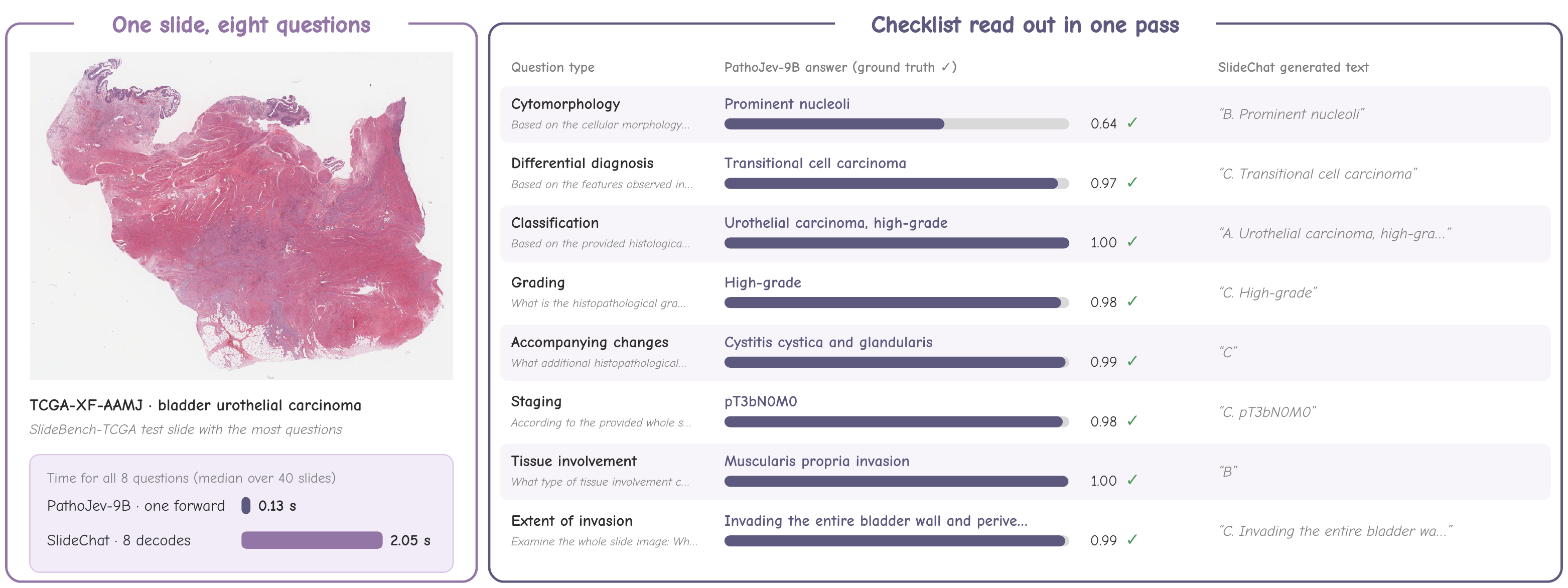}
    \caption{\textbf{Eight benchmark questions over one slide.} PathoJev-9B answers all eight SlideBench-TCGA questions for TCGA-XF-AAMJ through a batched question evaluation after shared-prefix prefill. Bars show the probability assigned to each predicted answer; the right column shows SlideChat outputs. These are the benchmark's question set, including its staging item. The displayed 0.13\,s and 2.05\,s are median times over 40 slides at $N=8$.}
    \label{fig:qual_multiquestion}
\end{figure*}

Figures~\ref{fig:qual_multiquestion}--\ref{fig:qual_scale} illustrate multi-question slide assessment, image-conditioned answers, and probability readout on WSI and ROI inputs. Figure~\ref{fig:qual_same_question} additionally reports predictions over all 40 slides asked the same grading question.

\subsection{Case Selection and Image Display}
\label{app:qual_selection}

\paragraph{Selection rules.}
Figure~\ref{fig:qual_multiquestion} uses the SlideBench-TCGA test slide with the largest number of questions, TCGA-XF-AAMJ, which has eight. For Figure~\ref{fig:qual_same_question}, we use the WSI-Bench grading question whose wording and options are identical across 40 slides. Within each ground-truth grade, we shuffle cases with seed 0 and select the first case answered correctly by PathoJev-9B.

\begin{figure*}[tp]
    \centering
    \includegraphics[width=\textwidth]{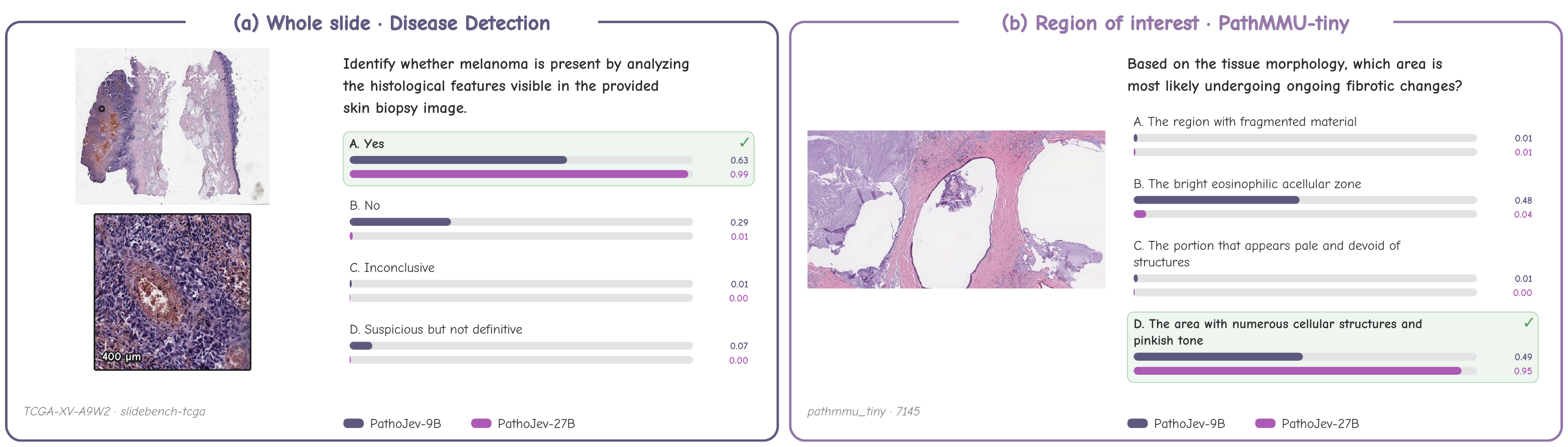}
    \caption{\textbf{Confidence differences between model scales.} Both models answer each question correctly, with examples selected for a top-1 probability increase of at least 0.2 from 9B to 27B. The displayed probabilities increase from 0.63 to 0.99 on the WSI example and from 0.49 to 0.95 on the ROI example.}
    \label{fig:qual_scale}
\end{figure*}

For Figures~\ref{fig:qual_probabilities}--\ref{fig:qual_scale}, candidates have at most five options and question text of at most 200 characters, extended to 400 for MedXpertQA. We exclude staging, TNM, size, and negative-form questions containing ``not,'' ``least,'' or ``except.'' A lexical filter excludes questions containing a content word of at least five letters from the correct answer, matching the first six letters. Each example appears in one figure. Eligible question IDs are ordered by the SHA-1 hash of \texttt{seed0:} concatenated with the ID, and the first eligible case is used for each panel.

Figure~\ref{fig:qual_probabilities} selects WSI questions answered correctly by PathoJev-9B and incorrectly by SlideChat, with top-1 probability above 0.95 in panel (a) and between 0.5 and 0.8 in panel (b). Figure~\ref{fig:qual_roi} selects one microscopy example per ROI benchmark that both PathoJev scales answer correctly. Visual filtering excludes three PathVQA gross or clinical photographs and one Quilt-VQA frame showing the speaker, with the exclusions recorded in the selection script. Figure~\ref{fig:qual_scale} selects questions both scales answer correctly and for which the 27B top-1 probability exceeds the 9B probability by at least 0.2.

\newpage
\paragraph{Display crops.}
WSI insets in Figures~\ref{fig:qual_same_question}, \ref{fig:qual_probabilities}, and \ref{fig:qual_scale} show $400\,\mu\mathrm{m}$ fields from IDC slide pyramids. They use the $10\times$ level (approximately $1\,\mu\mathrm{m}$/pixel), except TCGA-EM-A3SY, which uses the $5\times$ level (approximately $2\,\mu\mathrm{m}$/pixel). Windows are ranked on the thumbnail tissue mask by mean $255-R$, a hematoxylin proxy, with at least 90\% tissue coverage and background, pen marks, and very dark pixels excluded. Among up to ten non-overlapping windows in descending order, we use the first with fewer than 0.5\% pixels below grayscale intensity 60. Black boxes mark the displayed fields. These crops are for visualization; WSI inference uses the whole-slide grid. Screen-recorded ROI frames are cropped to the slide viewer for display.

\subsection{Example Predictions}
\label{app:qual_predictions}

\begin{figure*}[tp]
    \centering
    \includegraphics[width=\textwidth]{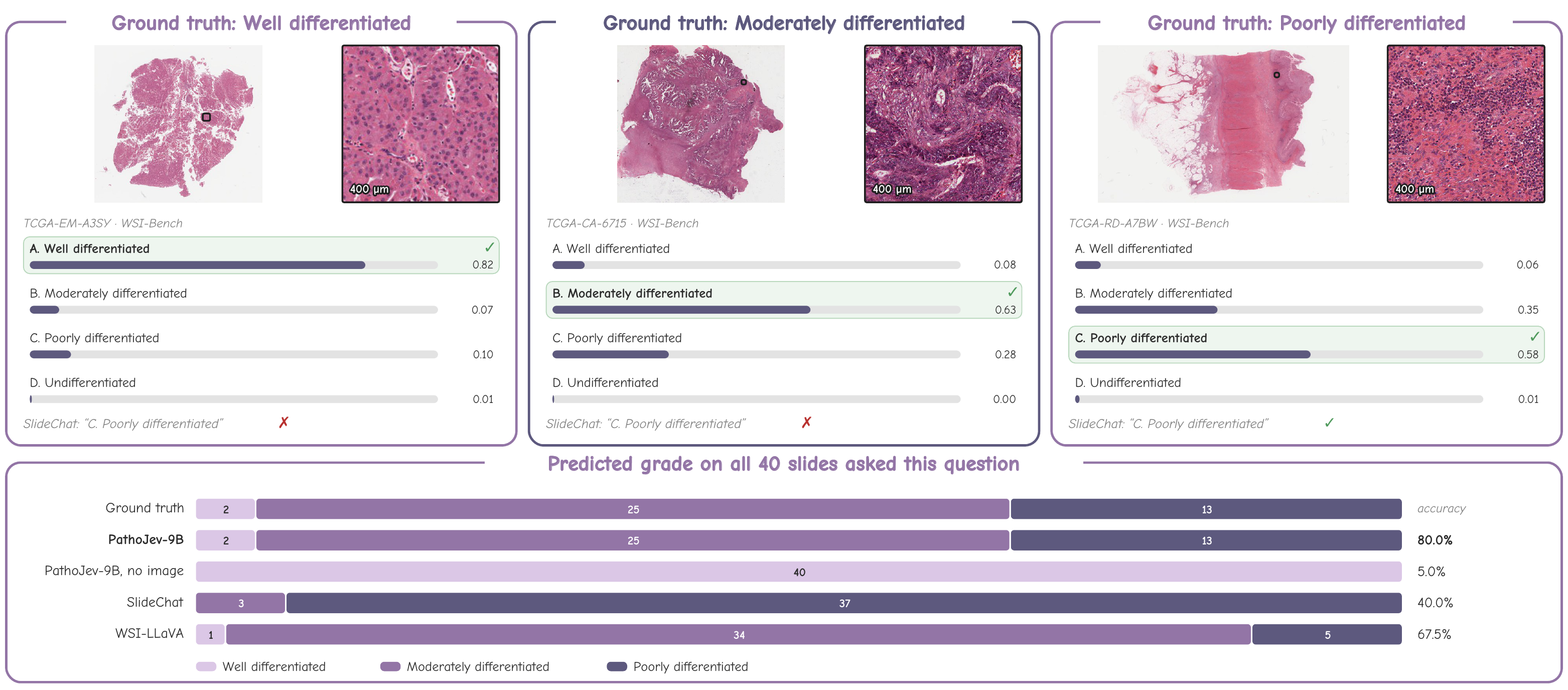}
    \caption{\textbf{The same question across different slides.} Top: one correctly answered example per ground-truth grade, selected with seed 0. Bottom: predicted grade counts and question-level accuracy over all 40 WSI-Bench slides asked this question. PathoJev-9B achieves 80.0\% accuracy with images; without images it predicts well differentiated for all slides and achieves 5.0\%. Insets show $400\,\mu\mathrm{m}$ fields for display; the model receives the whole-slide grid.}
    \label{fig:qual_same_question}
\end{figure*}

\begin{figure*}[tp]
    \centering
    \includegraphics[width=\textwidth]{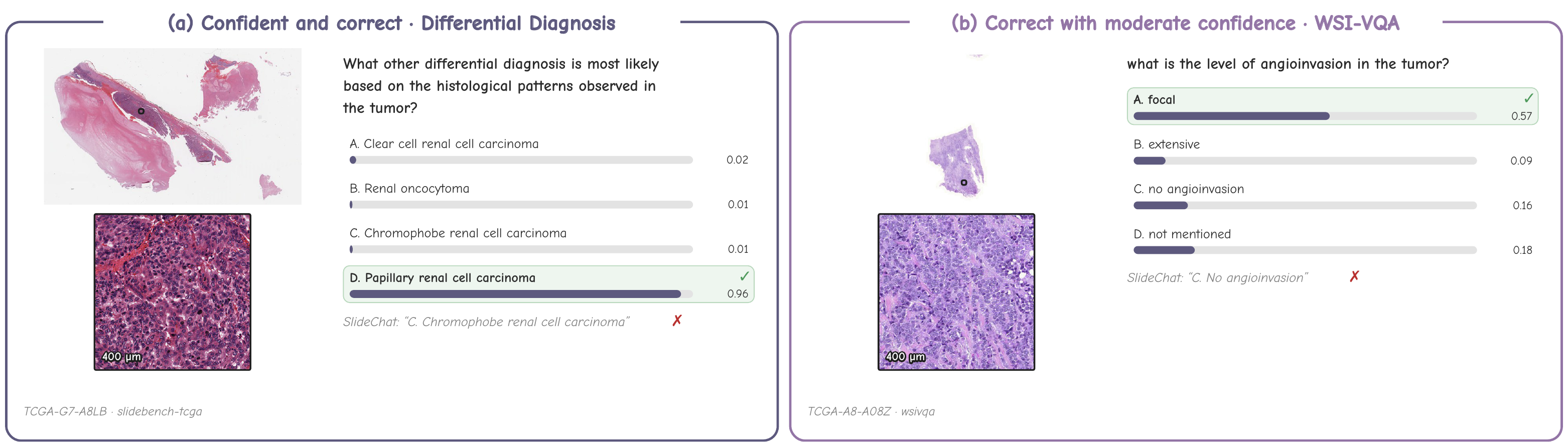}
    \caption{\textbf{Answer distributions for whole-slide decisions.} Examples answered correctly by PathoJev-9B and incorrectly by SlideChat. (a) The model assigns 0.96 probability to papillary renal cell carcinoma. (b) It assigns 0.57 to focal angioinvasion, with the remaining probability distributed across the other options. Selection and display crops follow Appendix~\ref{app:qual_selection}.}
    \label{fig:qual_probabilities}
\end{figure*}

\begin{figure*}[tp]
    \centering
    \includegraphics[width=\textwidth]{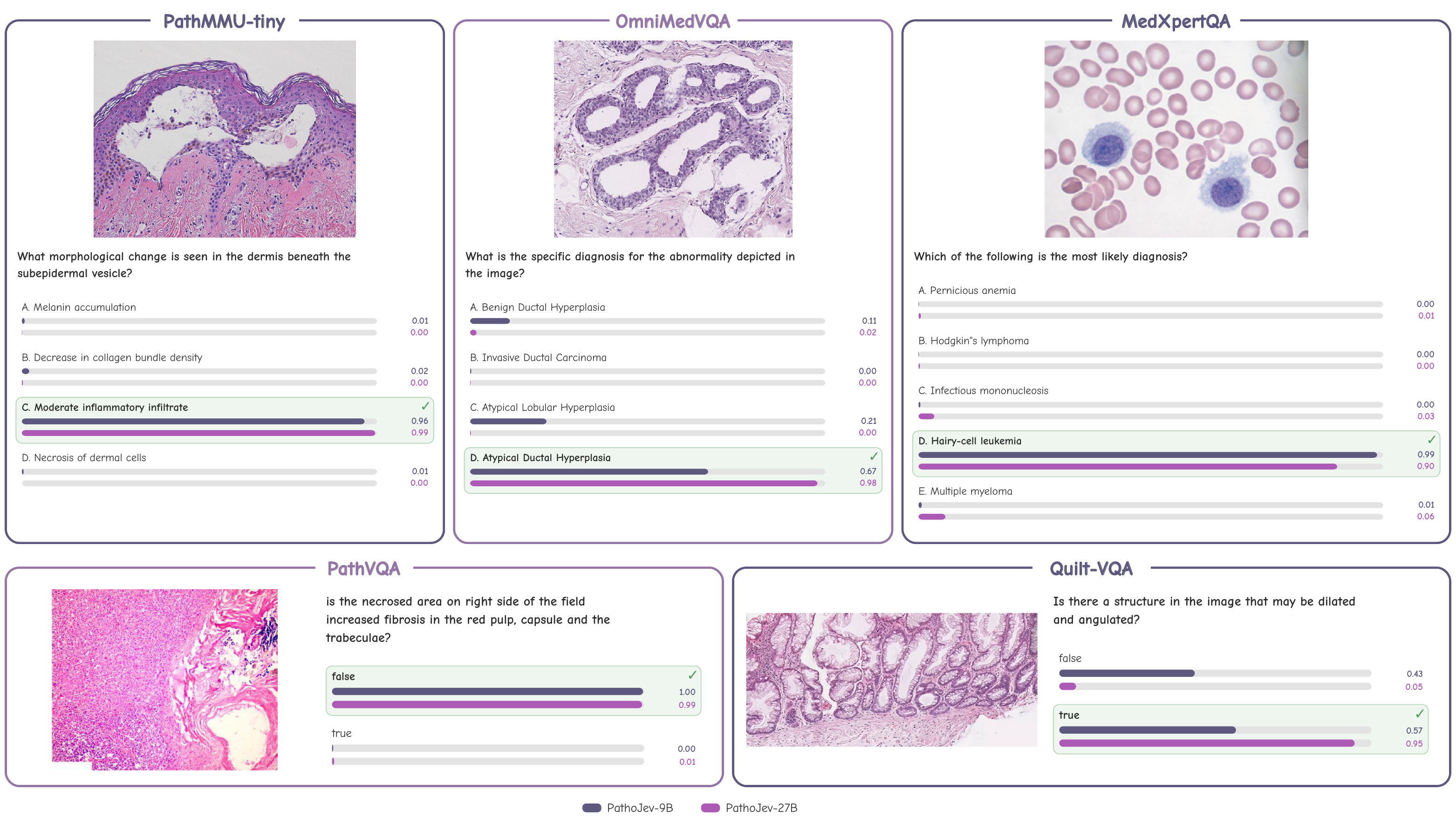}
    \caption{\textbf{Probability readout across five ROI benchmarks.} One microscopy question per benchmark answered correctly by both PathoJev-9B and PathoJev-27B. Paired bars show each model's probabilities over the supplied alternatives. Screen-recorded frames are cropped to the slide viewer for display.}
    \label{fig:qual_roi}
\end{figure*}

PathoJev-9B answers all eight benchmark questions in Figure~\ref{fig:qual_multiquestion} correctly through batched readout over a shared slide prefix. Figure~\ref{fig:qual_same_question} isolates a repeated question across different images: accuracy over the 40 slides is 80.0\% with images and 5.0\% without them, compared with 40.0\% for SlideChat and 67.5\% for WSI-LLaVA. Figures~\ref{fig:qual_probabilities} and~\ref{fig:qual_roi} show the full candidate distributions for WSI and ROI examples.

\onecolumn
\twocolumn
\section{Limitations and Future Work}
\label{app:future_work}

\smallskip\noindent\textbf{Extending the decision space.}
PathoJev currently predicts categorical and binary answers to questions with supplied candidates. Its evaluation focuses on judgments over individual slides and ROI images. Extending the interface to continuous measurements and spatial candidates would connect these judgments to quantitative and localized tissue assessment. Such outputs would require corresponding supervision and decision rules, while retaining the separation between shared visual context and question-specific computation.

\smallskip\noindent\textbf{Composing pathology workflows.}
A higher-level pathology assistant could formulate questions from a reporting goal, propose answer candidates, and assemble PathoJev's outputs into a structured account of the tissue findings. The assistant would manage question selection and report synthesis, while PathoJev would evaluate each visual judgment against the cached slide context. Answer probabilities could help prioritize the next question or select a region for closer inspection. This direction extends the use of the decision interface from a predefined question set to an evolving sequence of pathology queries.

\smallskip\noindent\textbf{Connecting multiple tissue views.}
Another direction is to combine decisions across slides, stains, and magnifications. Maintaining a separate visual cache for each image could support repeated queries within a view, while a coordinating model selects relevant views and combines their findings. Associating each decision with its source image would preserve the connection between specimen-level summaries and the underlying tissue evidence.


\end{document}